\documentclass{article} 
\usepackage[preprint]{colm2026_conference}
\usepackage{hyperref}       
\usepackage[T1]{fontenc}
\hypersetup{
	colorlinks=true,
	linkcolor=red,
	filecolor=blue,      
	urlcolor=magenta,
	citecolor={blue!50!green},
}
\usepackage{url}            
\usepackage{booktabs}       
\usepackage{microtype}      
\usepackage{xcolor}         
\usepackage{latexsym}
\usepackage{inconsolata}

\usepackage{graphicx}
\usepackage{subcaption}
\usepackage{multirow}
\usepackage{tabularx}
\usepackage{amsmath}
\usepackage{amssymb}
\usepackage{wrapfig}
\usepackage{xcolor}
\usepackage{colortbl}
\usepackage{array}
\definecolor{Gray}{gray}{0.9}
\usetikzlibrary{arrows.meta,positioning,fit,shapes.geometric,calc}
\usepackage[edges]{forest}
\newcommand{\thcell}[1]{\rule{0pt}{2.75ex}#1\rule[-1.05ex]{0pt}{0pt}}
\newcommand{\theadrow}{\rowcolor{Gray}}
\newcommand{\gtoprule}{\specialrule{\heavyrulewidth}{0pt}{0pt}}
\newcommand{\gmidrule}{%
  \noalign{\begingroup\color{Gray}\hrule height 1.25pt\endgroup}%
  \specialrule{\lightrulewidth}{0pt}{0pt}}
\newcommand{\gbottomrule}{\specialrule{\heavyrulewidth}{0pt}{0pt}}
\definecolor{blizzardblue}{rgb}{0.67, 0.9, 0.93}
\definecolor{aliceblue}{rgb}{0.94, 0.97, 1.0}
\definecolor{babyblueeyes}{rgb}{0.63, 0.79, 0.95}
\definecolor{beaublue}{rgb}{0.74, 0.83, 0.9}
\definecolor{azure(web)(azuremist)}{rgb}{0.94, 1.0, 1.0}
\usepackage{mdframed}
\usepackage{algorithm}
\usepackage{algpseudocode}
\usepackage[most]{tcolorbox}
\definecolor{lightgray}{gray}{0.9}
\definecolor{lightred}{rgb}{1,0.9,0.9}
\definecolor{lilac_light}{HTML}{C596B0}
\definecolor{lilac_dark}{HTML}{76415E}
\usepackage{caption}
\usepackage{fancyhdr}
\usepackage{etoc}

\algnewcommand\Break{\State \textbf{break}}
\algnewcommand\Comment[1]{\hfill $\triangleright$ \textit{\small #1}}
\definecolor{lightgray}{gray}{0.95}
\definecolor{deepblue}{RGB}{70,130,180}
\definecolor{deepgray}{RGB}{119,136,153}
\definecolor{opcolor}{RGB}{194,98,30} 
\lstdefinestyle{prompt}{
    basicstyle=\ttfamily\fontsize{7pt}{8pt}\selectfont,
    frame=none,
    breaklines=true,
    backgroundcolor=\color{lightgray},
    breakatwhitespace=true,
    breakindent=0pt,
    escapeinside={(*@}{@*)},
    numbers=none,
    numbersep=5pt,
    xleftmargin=5pt,
    aboveskip=2pt,
    belowskip=2pt,
}
\tcbset{
  aibox/.style={
    top=10pt,
    colback=white,
    enhanced,
    center,
  }
}
\newtcolorbox{AIbox}[2][]{aibox,title={#2},#1}
\newtcolorbox{questionbox}[1][]{%
  enhanced,
  width=\linewidth,
  colback=teal!20,     
  colframe=white,      
  boxrule=0pt,
  arc=0pt, outer arc=0pt,  
  left=8pt, right=8pt,
  top=6pt, bottom=6pt,
  before skip=6pt, after skip=6pt, 
  borderline west={3pt}{0pt}{blue!50!green}, 
  #1
}
\newcolumntype{Y}{>{\raggedright\arraybackslash}X}

\title{%
  DeepRefine: Agentic Knowledge Refinement via Reinforcement Learning%
}

\author{%
  Haoyu Huang$^{1}$\thanks{Contact: hhuangcp@connect.ust.hk}, \quad Jiaxin Bai$^{2}$\thanks{Corresponding Author, baijiaxin@hkbu.edu.hk},\quad Shujie Liu$^{3}$,\quad Yang Wei$^{1}$,\quad Huihao Jing$^{1,4}$, \\\textbf{Hong Ting Tsang}$^{1}$, \textbf{Yisen Gao$^{1}$,\quad Zhongwei Xie$^{1}$,\quad Yufei Li$^{1}$,\quad Yangqiu Song$^{1}$} \\
  $^{1}$HKUST,\quad $^{2}$HKBU,\quad $^{3}$Microsoft Research Asia,\quad $^{4}$MODEIO.AI\\
  Hong Kong, SAR, China \\
  \href{https://github.com/HKUST-KnowComp/DeepRefine}
  {
      \raisebox{-0.1ex}{\includegraphics[height=1.0em]{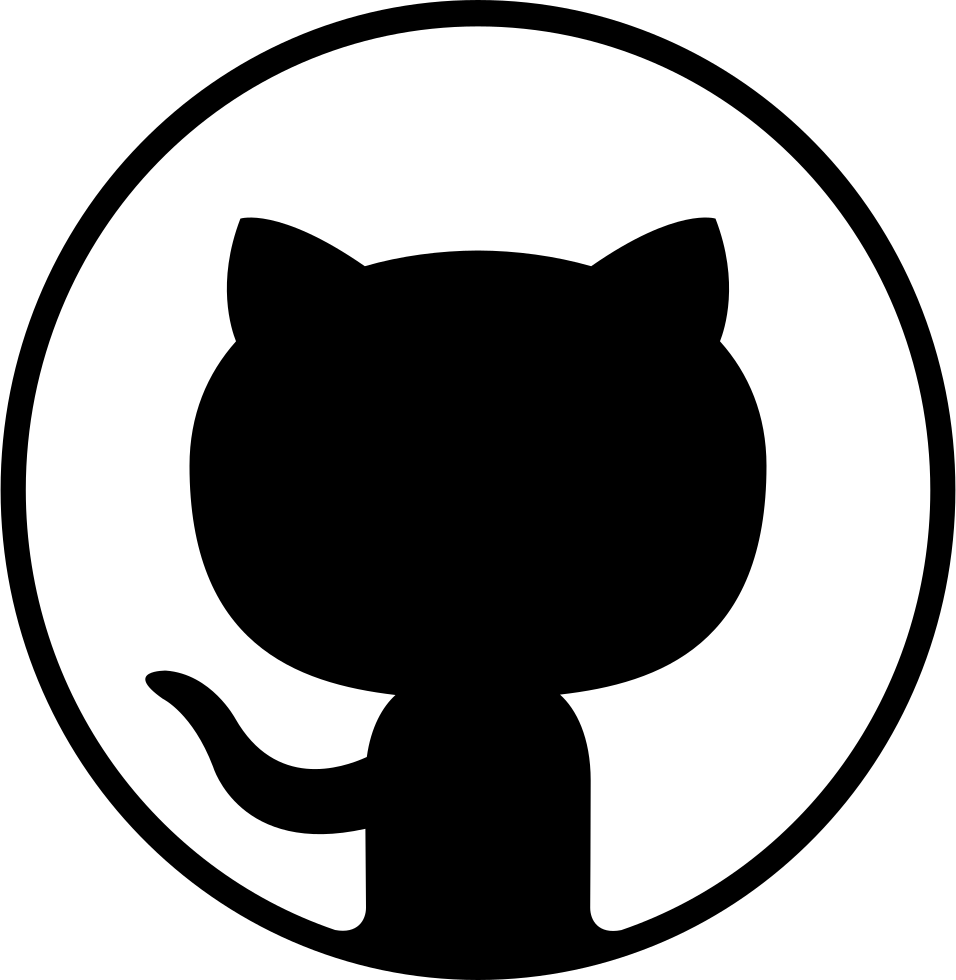}~Project}
  } \quad 
  \href{https://huggingface.co/collections/HaoyuHuang2/deeprefine}{%
    \raisebox{-0.1ex}{\includegraphics[height=1.0em]{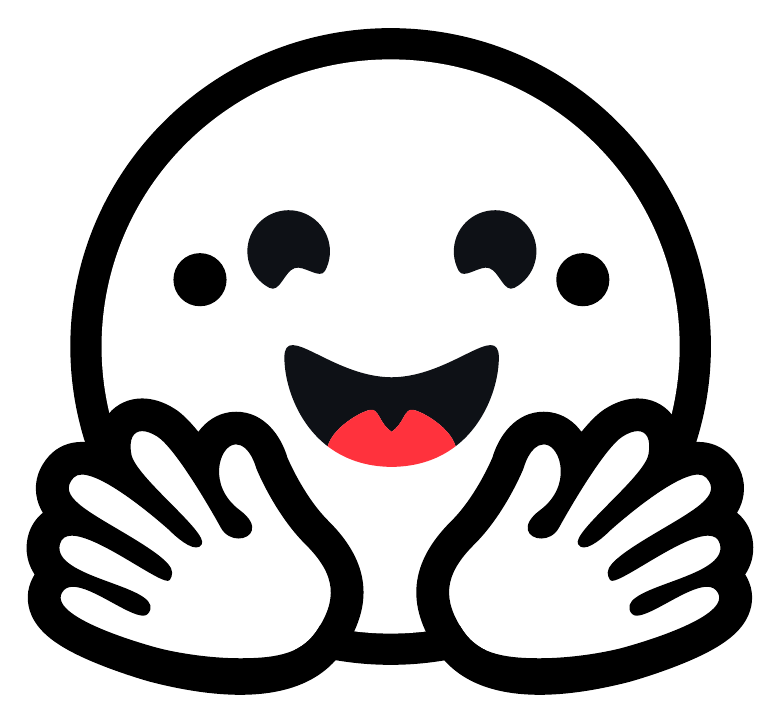}~Models}
  }
}

\begin{document}

\maketitle
\etocdepthtag.toc{mtchapter}
\etocsettagdepth{mtchapter}{none}
\etocsettagdepth{mtappendix}{none}
\begin{abstract}
External knowledge enables large language model (LLM) agents to ground their actions and decisions beyond intrinsic parametric memory in open-ended, knowledge-intensive downstream tasks. Yet the quality of the underlying knowledge bases is systematically limited by incompleteness, incorrectness, or redundancy, manifested as missing evidence or cross-document links, low-confidence or imprecise claims, and ambiguous or coreference resolution issues. Such defects compound under iterative use, degrading retrieval fidelity and downstream task performance. We present \textbf{DeepRefine}, a reinforcement learning framework for agentic knowledge refinement that evolves the quality of any pre-constructed structured knowledge bases, e.g., knowledge graphs or LLM-Wikis, with user queries to make it more suitable for the downstream tasks. DeepRefine performs multi-turn interactions with the knowledge base and conducts abductive diagnosis over the interaction history, localizes likely defects, and executes targeted refinement actions for incremental knowledge base updates. To further optimize refinement policies of DeepRefine without golden refinement trajectories, we introduce a Gain-Beyond-Draft (GBD) reward and train the reasoning process end-to-end via reinforcement learning. Extensive experiments demonstrate consistent downstream gains over strong baselines.
\end{abstract}

 \section{Introduction}\label{introduction}
 Large Language Models (LLMs) have shown remarkable capabilities in handling natural language understanding and generation tasks~\citep{yang2025qwen3, jaech2024openai}. And retrieval-augmented generation (RAG) grounds LLMs in external knowledge bases (KBs) outside model parameters~\citep{lewis2020retrieval, gao2023retrieval}. A dominant recipe indexes external knowledge as vectors for dense retrieval. Graph-based RAG systems additionally organize salient content into graph-structured KBs to support multi-hop retrieval and reasoning~\citep{edge2024local, guo2024lightrag, gutierrez2025rag, huang2025retrieval, yang2026graph}. In both cases, the external KB is usually built once and static at serving time, which provides LLMs with richer external knowledge and enhances their capabilities in domain-specific or knowledge-intensive tasks.
 
\begin{wrapfigure}{R}{0.48\textwidth}
   \centering
   \vspace{-10pt}
   \includegraphics[width=\linewidth]{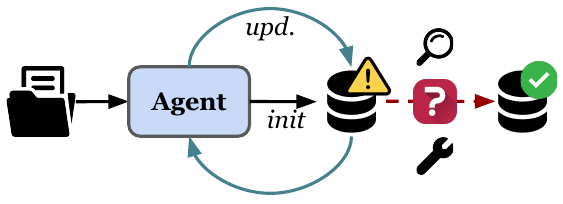}
   \caption{The quality problems of agentic KBs (left) and the challenges of KB refinement (right).}
   \label{fig:challenge}
   \vspace{-6pt}
 \end{wrapfigure}

 Yet a static KB is largely stateless with respect to ongoing use. It does not accumulate structured lessons from repeated queries, user-specific context, or iterative agent workflows. Recently some agentic KBs (e.g., LLM-Wiki~\citep{karpathy2026llmwikigist}) are proposed to address this gap by having LLM-based agents ingest sources, synthesize articles, and maintain cross-references over time. For example, toolkits that compile documents using agentic workflows and file outputs back into a user-owned persistent store. As shown in the left part of Figure~\ref{fig:challenge}, such systems make external KBs evolvable rather than frozen at test time.
 
 Despite this progress, both of the static and agentic KBs often exhibit systematic quality defects, including incompleteness, incorrectness, and redundancy~\citep{subagdja2024machine,10.3233/SW-160218}. In practice, these defects manifest as missing evidence chains or cross-document links, low-confidence or imprecise assertions, and ambiguous or coreference resolution issues. Such issues compound and degrade both retrieval fidelity and end-task performance. Recent studies mainly optimize the construction policy of KB builders~\citep{wang2025mem,tsang2025autograph}, however, in many real deployments the KB has already been constructed, so re-training the constructor and re-compiling the whole KB is often expensive and operationally impractical at scale~\citep{choubey2024distill,wolff2026cost}, motivating a post-construction knowledge refinement paradigm.
 
 Achieving effective knowledge refinement presents two core challenges: \textbf{(1) defect localization}, i.e., identifying problematic regions in a large KB without performing expensive traversal checks; and \textbf{(2) policy optimization without golden refinement trajectories}, i.e., even though ground truth answers are usually available in various benchmarks, it is still challenging for a model to learn effective refinement strategies with the absence of golden refinement trajectories or golden KBs.
 
 As shown in the right part of Figure~\ref{fig:challenge}, Challenge (1) arises because KBs are typically large and relationally connected, making exhaustive inspection infeasible for locating defects that stem from incompleteness, incorrectness, or redundancy~\citep{dong2025refining}. Challenge (2) arises because the LLM-based agents have difficulty refining KBs and learning how to refine KBs. The real-world refinement lacks golden trajectories and we generally do not observe the ``correct'' sequence of refinement actions that best aligns a KB with downstream tasks. Consequently, learning effective policies requires training objectives that can evaluate refinement quality through downstream task utility signals.
 
 To address these challenges, we propose \textbf{DeepRefine}, a reinforcement learning (RL) framework for agentic knowledge refinement that improves any pre-constructed KB, e.g., LLM-Wiki and knowledge graphs (KGs), with user queries to make it more suitable for the downstream tasks. DeepRefine follows a three-step reasoning process: \textbf{Answerability Judgement Loop}, \textbf{Error Abduction}, and \textbf{Refinement Actions Generation}. For Challenge (1), instead of traversing the full KB, DeepRefine performs multi-turn query-conditioned interaction with external knowledge to localize potentially defective neighborhoods. This agentic design substantially reduces refinement complexity while preserving effectiveness. For Challenge (2), DeepRefine abductively infers likely defect causes from interaction histories, then generates refinement actions to update the KB incrementally. Because the utility of refinement actions is only observable through downstream task performance and is non-differentiable, we introduce a \textbf{Gain-Beyond-Draft (GBD) reward} and optimize the refinement policy end-to-end via reinforcement learning.
 
 In summary, we make the following main contributions:
 
 \begin{itemize}
   \item To the best of our knowledge, \textbf{DeepRefine} is the first RL framework for agentic knowledge refinement, capable of evolving the quality of any pre-constructed structured knowledge base using only user queries to make it more suitable for the downstream tasks.
   \item We design a three-step reasoning process including \textbf{Answerability Judgement Loop}, \textbf{Error Abduction}, and \textbf{Refinement Actions Generation}. We further introduce a \textbf{Gain-Beyond-Draft (GBD) reward} with reinforcement learning to enhance the knowledge base refinement capabilities of LLMs.
   \item We conduct extensive experiments across five datasets to demonstrate that DeepRefine can steadily improve the performance of state-of-the-art baseline methods.
 \end{itemize}
 
 \section{Related Work}
 \subsection{Agentic Knowledge Base Management}
 KBs are effective for agentic use because they usually capture relational and multi-hop dependencies, organize evidence hierarchically~\citep{huang2025retrieval, zhang2025leanrag, xie2026survey}, and support efficient retrieval~\citep{yang2026graph}. Different from graph-enhanced RAG methods that largely treat constructed structures as static artifacts~\citep{edge2024local,jimenez2024hipporag}, recent agent memory systems (e.g., Zep~\citep{rasmussen2025zep}, Mem0~\citep{chhikara2025mem0}) emphasize continuous maintenance and evolution for new knowledge. Existing RL-based approaches, such as Mem-$\alpha$~\citep{wang2025mem}, Memory-R1~\citep{yan2025memory}, and AutoGraph-R1~\citep{tsang2025autograph}, primarily optimize construction-time KB management policies. In contrast, our work focuses on post-construction refinement of already deployed KBs.
 
 \subsection{LLMs and Reinforcement Learning}
 Reinforcement learning (RL)~\citep{kaelbling1996reinforcement} is an effective optimization framework for improving the sequential decision-making capabilities of LLMs through learning from environmental interaction and reward feedback~\citep{xi2025survey}. Reinforcement Learning from Human Feedback (RLHF)~\citep{ouyang2022training} is one of the foundational methods used to align LLM outputs with preferences. And Proximal Policy Optimization (PPO)~\citep{schulman2017proximal}, and Group Relative Policy Optimization (GRPO)~\citep{shao2024deepseekmath} have provided scalable and efficient optimization algorithms for various structured decision-making tasks in several domains. For example, Graph-R1~\citep{luo2025graph} employs RL to learn effective graph-structured tools navigation to retrieve more accurately. Search-R1~\citep{jin2025search} trains LLMs to optimize web search queries using RL to maximize final answer correctness and \texttt{s3}~\citep{jiang2025s3} further proposes a principled, model-agnostic reward signal that quantifies improvements over standard retrieval.
 
 \subsection{Knowledge Refinement}
 Knowledge refinement aims to mitigate incompleteness, incorrectness, and redundancy in existing artifacts~\citep{subagdja2024machine,10.3233/SW-160218}. Classical KG completion is one of the knowledge refinement methods, such as KG-BERT~\citep{yao2019kg} and KGRefiner~\citep{saeedizade2022kgrefiner}, are typically limited to adding links or relations. Entity alignment and disambiguation methods, such as NeuSymEA~\citep{chen2024neuro} and K-NED~\citep{feng2020knowledge}, focus on correction and simplification. In contrast, knowledge refinement actually has a broader scope that subsumes both types of operations~\citep{10.3233/SW-160218}, yet remains underexplored. TRAIL~\citep{zhao2025trail} enables LLM-based agents to iteratively explore and update KGs during reasoning, however, its edits are not explicitly optimized against downstream task utility. Our method instead learns refinement policies with an explicit downstream reward, yielding tighter alignment between KB updates and end-task performance.
 
 \section{Preliminaries}\label{sec:preliminaries}
 In this section, we formalize the KB and the refinement task with respect to RAG-based downstream tasks. We use constructed KGs and LLM-Wikis as the KBs. They are represented as structured triples where the heads and tails may be entities, events, or document-grounded spans, and relations may capture taxonomy, time, causality, or cross-document linkage.
 
 We denote the pre-constructed KB by a set of triples $\mathcal{G}_{\text{f}} = \{(h, r, t) \mid h, t \in \mathcal{E}_{\text{f}},\, r \in \mathcal{R}_{\text{f}}\}$, where $\mathcal{E}_{\text{f}}$ is the set of triples and $\mathcal{R}_{\text{f}}$ is the set of relations. The KBs (KGs or LLM-Wikis) may be produced by any LLM-based constructors. We are also given a set of user queries $\mathcal{Q} = \{q_1, q_2, \ldots, q_N\}$ related to the $\mathcal{G}_{\text{f}}$. The goal of agentic knowledge refinement is to update $\mathcal{G}_{\text{f}}$ with user queries in $\mathcal{Q}$ to make it more suitable for the downstream RAG tasks. In real-world applications, these queries can be the users' query history and our method aims to make use of them to evolve the quality of the pre-constructed KB.
 
 Then we further formulate KB refinement as generating sequences of actions that mitigate incompleteness, incorrectness, and redundancy in $\mathcal{G}_{\text{f}}$. This process can be represented as:
 \begin{equation}
   \underset{S_{\mathcal{A}_{\text{q}}}\in S}{\arg\max}~p_{\theta}(S \mid \mathcal{G}_{\text{f}}, \mathcal{Q}) = \{\mathcal{A}_{\text{q}}^{1}, \ldots, \mathcal{A}_{\text{q}}^{|\mathcal{Q}|}\},
 \end{equation}
 where $S_{\mathcal{A}_{\text{q}}}$ is the set of action sequence $\mathcal{A}_{\text{q}}^{i} = \{a_1, a_2, \ldots, a_{M_{i}}\}$ with respect to the $i$-th query $q \in \mathcal{Q}$, $S$ is the possible output set, and $p_{\theta}$ is the policy model parameterized by $\theta$. These actions are then applied to improve the quality of the KB $\mathcal{G}_{\text{f}}$.
 
 \section{Methodology}\label{sec:methodology}
 \begin{figure*}[t]
   \centering
   \includegraphics[width=0.95\linewidth]{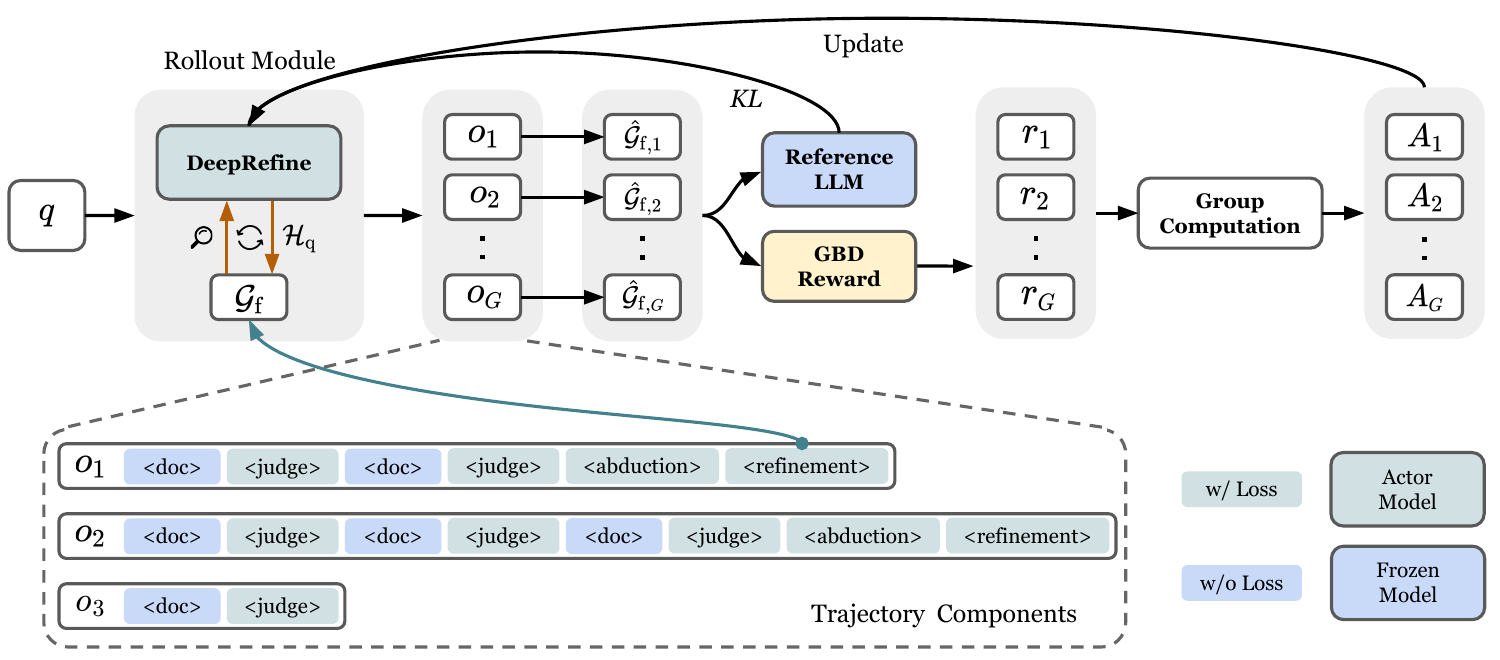}
   \caption{Demonstration of GRPO training for the agentic knowledge refinement (DeepRefine), where the \textcolor{orange}{orange} arrows denote the multi-turn interactions of DeepRefine with KBs and the \textcolor{teal}{green} arrows denote the refinement actions on KBs.}
   \label{fig:overview}
 \end{figure*}
 \vspace{-0.25cm}
 \subsection{Reasoning Steps Design}\label{subsec:reasoning_steps_design}
 
 In this section, we will demonstrate the reasoning steps of DeepRefine for the agentic knowledge refinement. Note that it is a multi-turn reasoning process, which means the preceding reasoning steps will be served as the context for the subsequent reasoning steps.
 
 \textit{\textbf{Answerable Judgement Loop.}} Initially, DeepRefine will judge the answerability of the given query with several interactions with the KB $\mathcal{G}_{\text{f}}$, as illustrated in the loop of the rollout module in Figure~\ref{fig:overview}. If the query is answerable in the first interaction, DeepRefine will directly skip that. Otherwise, DeepRefine will refine $\mathcal{G}_{\text{f}}$ in later reasoning steps to make it more easily to be answered. In the first interaction with $\mathcal{G}_{\text{f}}$, as shown in Eq.~\ref{eq:top_k_subkg}, DeepRefine will conduct a simple dense retrieval to obtain the top-$N$ related triples to the query with their embeddings, which can be considered as a $0$-hop retrieved subgraph $\mathcal{G}^{(0)}_{\text{q}}$.
 \begin{equation}
 \mathcal{G}^{(0)}_{\text{q}} = \text{Top-}k(q, \mathcal{G}_{\text{f}}, N).
 \label{eq:top_k_subkg}
 \end{equation}
 Then for each of the following interactions, DeepRefine will continue to expand the retrieved subgraph if the answerability is not satisfied in the previous interaction. And only the newly retrieved triples will be added as user messages in the next turn's assistant messages generation (judgement). To be specific, the subgraph expanding procedure is defined as follows:
 \begin{equation}
 \mathcal{G}_{\text{cand}}^{(i)} = \{(h, r, t) \in \mathcal{G}_{\text{f}} \mid h \text{ or } t \in \mathcal{E}^{(i-1)}_{\text{q}} \},
 \label{eq:collect_neighbors}
 \end{equation}
 \begin{equation}
 \mathcal{G}_{\text{pruned}}^{(i)} = \text{Top-}k(q, \mathcal{G}_{\text{cand}}^{(i)}, M),
 \label{eq:top_m}
 \end{equation}
 \begin{equation}
 \mathcal{G}^{(i)}_{\text{q}} = \mathcal{G}^{(i-1)}_{\text{q}} \cup \mathcal{G}_{\text{pruned}}^{(i)},
 \label{eq:merge_subgraph}
 \end{equation}
 where $\mathcal{G}^{(i)}_{\text{cand}}$ is a set of new candidate triples that connect to the knowledge items in the subgraph $\mathcal{G}^{(i-1)}_{\text{q}}$ and $\mathcal{E}^{(i-1)}_{\text{q}}$ is the knowledge item set of the subgraph $\mathcal{G}^{(i-1)}_{\text{q}}$. To control the size of the retrieved subgraph with respect to the query $q$, we further prune the candidate triples with their embeddings, and select the top-$M$ related triples to the query, as shown in Eq.~\ref{eq:top_m}. Then we merge these new triples with the previous subgraph and obtain the $i$-hop retrieved subgraph $\mathcal{G}^{(i)}_{\text{q}}$. The process of interactions will continue until the answerability is satisfied or the maximum number of expansion hops is reached. The judgement output will be enclosed within $\textcolor{opcolor}{\texttt{<judge>}}$ and $\textcolor{opcolor}{\texttt{</judge>}}$ tags as the prompt template demonstrated in Figure~\ref{answerable_judgement_prompt} of Appendix~\ref{sec:prompt_templates}. After it terminates, the query-specific interaction history $\mathcal{H}_{\text{q}}$ with a total length of $L$ can be finally obtained for the following reasoning steps. The interaction history consists of the retrieved subgraph at each interaction step and the query:
 \begin{equation}
 \mathcal{H}_{\text{q}} = \bigcup_{i=0}^{L-1} \{(q, \mathcal{G}^{(i)}_{\text{q}}, \mathcal{J}^{(i)}_{\text{q}})\},
 \label{eq:interaction_history}
 \end{equation}
 where $\mathcal{J}^{(i)}_{\text{q}}$ is the answerable judgement result of the query $q$ at the $i$-th interaction step.

 \textit{\textbf{Error Abduction.}} We assume that there are some potential issues in the triples of a query-specific subgraph if the query is not answerable with the $0$-hop retrieved subgraph. In this reasoning step, given the interaction history $\mathcal{H}_{\text{q}}$, DeepRefine will abductively reason the potential issues $\mathcal{I}_{\text{q}}$ in the retrieved subgraphs from three perspectives: incompleteness, errors and redundancy, which are based on the definition of KG refinement~\citep{subagdja2024machine} proposed by \citet{10.3233/SW-160218}. The error reasons will be generated within the $\textcolor{opcolor}{\texttt{<abduction>}}$ and $\textcolor{opcolor}{\texttt{</abduction>}}$ tags as the prompt template demonstrated in Figure~\ref{error_abduction_prompt} of Appendix~\ref{sec:prompt_templates}. For example, if the query is about a multi-hop relationship between two entities, but the retrieved subgraphs wrongly include or even do not contain some intermediate entities or relations, DeepRefine will reason the incorrect or missing parts that may lead to the unanswerable result. Similarly, as for redundancy, DeepRefine will also check if there is any ambiguous knowledge that would affect the answerability. The analysed results in this step will be served as the reference for the following refinement actions generation.
 
 \textit{\textbf{Refinement Actions Generation.}} Based on the analysed potential issues $\mathcal{I}_{\text{q}}$ in the subgraphs retrieved from the previous interactions, DeepRefine will edit the full KB $\mathcal{G}_{\text{f}}$ accordingly to improve its quality. To reduce token costs and improve the refinement efficiency, instead of reconstructing the subgraphs and then inserting into the full KB, for each query $q$, DeepRefine will generate a series of refinement actions $\mathcal{A}_{\text{q}}$ within the $\textcolor{opcolor}{\texttt{<refinement>}}$ and $\textcolor{opcolor}{\texttt{</refinement>}}$ tags to edit the full KB directly:
 \begin{equation}
   \mathcal{A}_{\text{q}} = \underset{\mathcal{A}_{\text{q}} \in \mathcal{A}}{\arg\max}~ p_{\theta}(\mathcal{A} \mid \mathcal{G}_{\text{q}}^{(L)}, \mathcal{I}_{\text{q}}),
 \label{eq:action_gen}
 \end{equation}
 where $\mathcal{A}$ is the set of possible output from DeepRefine parameterized by $\theta$. The generated refinement actions will be parsed into three pre-defined types of operators: $\textcolor{opcolor}{\texttt{insert\_edge()}}$, $\textcolor{opcolor}{\texttt{delete\_edge()}}$, and $\textcolor{opcolor}{\texttt{replace\_node()}}$. The action space is designed to refine the KB along the three issues discussed above. The pre-defined operators are just interfaces, which makes this paradigm can be generalizably applied across various graph databases. We also further elaborate on the rationality of the action space design from both functionality and flexibility perspectives in Appendix~\ref{appendix:action_space}. After applying the actions, we will obtain the refined KB $\hat{\mathcal{G}_{\text{f}}}$. And the full rollout process is also elaborated in Algorithm~\ref{alg:reasoning}.
 
 \subsection{Refinement Policy Optimization}
 \textbf{\textit{Reward Design}}. To optimize the knowledge refinement policy $\pi_{\theta}$, we consider knowledge refinement as an end-to-end RL problem. Specifically, as shown in the Figure~\ref{fig:overview}, inspired by the prior work~\citep{jiang2025s3}, we introduce Gain Beyond Draft (GBD) as the reward objective, which quantifies the gain in RAG generation accuracy achieved by the refined KB $\hat{\mathcal{G}_{\text{f}}}$ relative to the draft KB $\mathcal{G}_{\text{f}}$:
 \begin{equation}
 \text{GBD}(q) = \text{F1}(A_{\text{refined}}, A) - \text{F1}(A_{\text{draft}}, A),
 \label{eq:gbd}
 \end{equation}
 where $A_{\text{refined}}$ and $A_{\text{draft}}$ are the generated answers with the refined KB $\hat{\mathcal{G}_{\text{f}}}$ and the draft one $\mathcal{G}_{\text{f}}$, which are both obtained through the subgraph retriever we mentioned in Section~\ref{sec:experimental_settings}. $A$ is the golden answer for the query and $\text{F1}(\cdot, \cdot)$ is the token-level F1 score.
 
 \textbf{\textit{Policy Optimization}}. To optimize the KB refinement policy $\pi_{\theta}$ of DeepRefine, we adopt the Group-Relative Policy Optimization (GRPO)~\citep{shao2024deepseekmath} algorithm to finetune the policy with the GBD reward.
 Formally, the objective is defined as:
 \begin{equation}
 \begin{aligned}
 &\mathcal{J}_{\text{GRPO}}(\theta) = \mathbb{E}_{\textbf{q} \sim P_Q, \{\textbf{o}_i\} \sim \pi_{\theta_{\text{old}}}(\cdot | \textbf{q})} \\
 &\left[ \frac{1}{G}\sum^{G}_{i=1}\frac{1}{|\textbf{o}_{i}|}\sum^{|\textbf{o}_{i}|}_{t=1}\text{min}\left( r_{i,t}(\theta)\hat{A}_{i,t}, \text{clip}\Bigl(r_{i,t}(\theta), 1 - \epsilon, 1 + \epsilon \Bigr)\hat{A}_{i,t} \right) - \beta \mathbb{D}_{\text{KL}}[\pi_{\theta} || \pi_{\text{ref}}] \right],
 \end{aligned}
 \end{equation}
 where $r_{i,t}(\theta) = \frac{\pi_{\theta}(o_{i,t}|\textbf{q}, \textbf{o}_{i, <t})}{\pi_{\theta_{\text{old}}}(o_{i,t}|\textbf{q}, \textbf{o}_{i, <t})}$ denotes the probability ratio and $\hat{A}_{i,t} = \frac{R_{i} - \mu_{R}}{\sigma_{R}}$ represents the  group-relative advantage for the entire refinement reasoning process. And $\pi_{\text{ref}}$ is the reference policy. During the training process, DeepRefine will generate the knowledge refinement actions together with the reasoning steps we described in Section~\ref{subsec:reasoning_steps_design} token by token, where $\textbf{o}_i = \{ o_{i,1}, \ldots, o_{i,T} \}$ represents the sequence of tokens forming the full reasoning output at the $i$-th group of $G$ samples. And the reward signal $R_i$ for the $i$-th query sample is calculated through the reward function, as illustrated in Eq.~\eqref{eq:gbd}. $\epsilon$ is a small clipping hyperparameter that ensures stable updates by preventing the new policy from straying too far from the old policy. The policy optimization process is also elaborated in Algorithm~\ref{alg:training}.
 \subsection{Inference}
 \begin{wrapfigure}{R}{0.45\textwidth}
   \centering 
   \includegraphics[width=\linewidth]{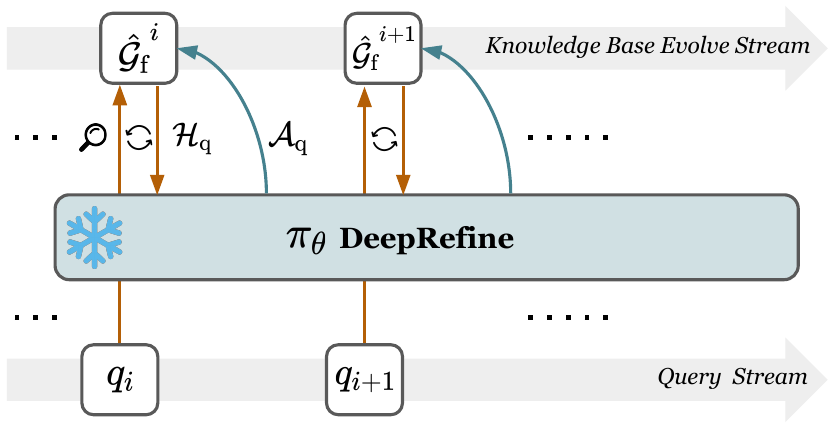}
   \caption{An overview of the inference process of DeepRefine.}
   \vspace{-0.35cm}
   \label{fig:golden-answer-free-inference}
   \end{wrapfigure}
 After the RL training, DeepRefine can be directly plugged into any pre-constructed structured KB. As shown in Figure~\ref{fig:golden-answer-free-inference}, during inference time, there could be two streams: one is the online query stream, the other one is the KB evolve stream. The first stream is intuitive, because users will sequentially ask the KB questions once it has been constructed. And the knowledge refinement process conducted by a frozen DeepRefine with respect to an online query will forma the second stream. Note that the refinement process can be asynchronous with the online service, which will not introduce any systematic latency. DeepRefine will process these queries sequentially with multi-step reasoning chains to improve the quality of the KB and make it more suitable for the downstream tasks.
 
 \section{Experiments}
 In this section, we will report the performance of DeepRefine on both RAG and long-term conversation memory benchmarks. We focus on the performance gain of DeepRefine over various baseline methods and the efficiency of the knowledge refinement process.
 
 \subsection{Experimental Settings}\label{sec:experimental_settings}
 \textbf{\textit{Implementation Details.}} We implement the DeepRefine training and inference processes using both of the \texttt{Qwen3-8B} and \texttt{Qwen3-4B-Instruct-2507en} as the backbone model for their better instruction following capabilities. And we employ the \texttt{Qwen3-Embedding-0.6B} as the embedding model for all vector search in both our method and the baseline methods and the frozen \texttt{Qwen2.5-7B-Instruct} as the reader model for the RAG generation. For both of the 4B and 8B models, we use a batch size of $64$ with a mini-batch size of $32$ and a group size of $16$ for the $40$ and $30$ steps' GRPO training, respectively. The learning rate is $1e$-$6$ and the micro-batch size is $4$. To further improve the efficiency of the refinement process, we select a subset of queries before refinement using a greedy maximum-coverage procedure over query-related triples, as described in Appendix~\ref{sec:coverage_based_query_selection}. Our models are trained on $8\times$ NVIDIA A800-SXM4-80GB GPUs. Results are averaged under 3 times.
 
 \textbf{\textit{Baselines.}} We select baselines from two perspectives. (1) We evaluate the performance gain from DeepRefine over various KB constructor: naive KB constructor, the RL fine-tuned constructor, which employs \textbf{AutoGraph-R1}~\citep{tsang2025autograph}, and is a stronger constructor, and the agentic knowledge constructor, which is a version of LLM-Wiki and employs \textbf{Graphify}~\citep{safishamsi2026graphify}. Both the base and RL fine-tuned constructors share the same base model \texttt{Qwen2.5-3B-Instruct}. And we uses the \texttt{Composer-1.5} of Cursor as the agent in Graphify. (2) Conditioned on each of the above constructor, we evaluate whether the performance gains brought by DeepRefine remain stable across various knowledge retrievers. For triple retrievers, we employ \textbf{Subgraph Retriever} and \textbf{ToG}~\citep{sun2024think}. For graph-based text retrievers, we employ \textbf{HippoRAG}~\citep{jimenez2024hipporag} and \textbf{HippoRAG2}~\citep{gutierrez2025rag}. We use the same settings of them as those in AutoGraph-R1.
 
 \textbf{\textit{Training Datasets.}} We randomly select $5,000$ samples from the training set of \textbf{HotpotQA}~\citep{yang2018hotpotqa} to construct the training data for DeepRefine, and split them into training and validation sets with a ratio of $8{:}2$. For each original HotpotQA data point, we construct an individual KB. During the rollout process of the training, DeepRefine only refines the KB associated with the query in each sample individually.
 
 \textbf{\textit{Evaluation Datasets.}} We evaluate DeepRefine on three types of benchmarks, which include \textbf{Simple QA}, \textbf{Multi-hop QA} and \textbf{Conversation QA}. For simple QA, we use the Natural Questions (\textbf{NQ})~\citep{adelani2021masakhaner} and \textbf{PopQA}~\citep{mallen2023not} datasets. For multi-hop QA, we use the \textbf{2WikiMultihopQA}~\citep{ho2020constructing} and \textbf{Musique}~\citep{trivedi2022musiquemultihopquestionssinglehop} datasets. For conversation QA, we use the \textbf{LOCOMO}~\citep{maharana2024evaluating} dataset. For NQ and PopQA, we adopt a unified knowledge corpus constructed from the lead sections of the December 2021 Wikipedia dump~\citep{izacard2023atlas}. For 2WikiMultihopQA and Musique, we construct benchmark-specific corpora from the documents associated with each dataset's $1,000$ evaluation instances, consistent with the settings of \citet{jimenez2024hipporag}. Similarly, for LOCOMO, we also sample $1,000$ evaluation instances from the dataset. Following existing works~\citep{trivedi2022musiquemultihopquestionssinglehop, jimenez2024hipporag, yan2025memory}, we employ the token-level F1 score as the metric for benchmark evaluation.
 
 \subsection{Experimental Results}
 \begin{table}[ht]
   \centering
   \caption{Averaged performance of our GBD reward finetuned DeepRefine on various knowledge retrievers. The table reports RAG F1 on five QA datasets, grouped by three constructors.}
   \resizebox{0.95\textwidth}{!}{%
   \begin{tabular}{>{\raggedright\arraybackslash}p{6.0cm}|ccccc|c}
   \gtoprule
   \theadrow
   & \multicolumn{2}{c}{\thcell{Simple QA}} & \multicolumn{2}{c}{\thcell{Multi-hop QA}} & \multicolumn{1}{c}{\thcell{Conversation}} & \multicolumn{1}{c}{\thcell{Overall}} \\
   \theadrow
   \multirow{-2}{*}{\thcell{Methods}} & \thcell{NQ*} & \thcell{PopQA*} & \thcell{2WikiQA} & \thcell{Musique} & \thcell{LOCOMO} & \thcell{Avg.} \\
   \midrule
   \multicolumn{7}{l}{\cellcolor{brown!20}\textit{\textbf{Naive Constructor}}} \\
   \midrule
    Subgraph & 26.43 & 54.48 & 33.53 & 13.82 & 25.86 & 30.82 \\
   \cellcolor{teal!10}Subgraph + \texttt{DeepRefine-4B} &\underline{29.55} &\underline{60.09} &\underline{36.04}&\underline{15.93} & \textbf{29.64}&\underline{34.25}\textcolor{red}{$\uparrow_{3.43}$} \\
   \cellcolor{teal!20}Subgraph + \texttt{DeepRefine-8B} &\textbf{30.36}
    &\textbf{60.37} &\textbf{38.39} &\textbf{16.18} & \underline{29.07}&\textbf{34.87}\textcolor{red}{$\uparrow_{4.05}$}\\
   \cline{1-7}
   ToG & 25.55 & 54.92 & 43.74 & 18.21 & 22.64 & 33.01 \\
   \cellcolor{teal!10}ToG + \texttt{DeepRefine-4B} &\underline{29.21} &\underline{59.27} &\underline{49.16} &\underline{25.73} &\underline{24.53} & \underline{37.58}\textcolor{red}{$\uparrow_{4.57}$}\\
   \cellcolor{teal!20}ToG + \texttt{DeepRefine-8B} &\textbf{30.22} &\textbf{59.48} &\textbf{51.15} &\textbf{27.24} &\textbf{24.96} &\textbf{38.61}\textcolor{red}{$\uparrow_{5.60}$} \\
   \cline{1-7}
    HippoRAG & \underline{35.54} & 63.56 & 49.77 & 26.54 & \textbf{33.25} & 41.73  \\
   \cellcolor{teal!10}HippoRAG + \texttt{DeepRefine-4B} & 35.51&\textbf{64.97}&\underline{50.52} &\underline{27.82} &32.53 &\underline{42.27}\textcolor{red}{$\uparrow_{0.54}$} \\
   \cellcolor{teal!20}HippoRAG + \texttt{DeepRefine-8B} &\textbf{36.35} &\underline{64.21} &\textbf{51.17} &\textbf{29.07} &\underline{33.24} &\textbf{42.81}\textcolor{red}{$\uparrow_{1.08}$} \\
   \cline{1-7}
   HippoRAG2 & 35.91 & 63.08 & 51.47 & 25.33 & 33.33 &41.82 \\
   \cellcolor{teal!10}HippoRAG2 + \texttt{DeepRefine-4B} &\underline{36.94} &\underline{64.65} &\underline{51.75} &\underline{26.79} &\underline{33.46}&\underline{42.72}\textcolor{red}{$\uparrow_{0.90}$} \\
   \cellcolor{teal!20}HippoRAG2 + \texttt{DeepRefine-8B} &\textbf{37.32} &\textbf{64.91} &\textbf{53.18} &\textbf{27.19} & \textbf{33.55}& \textbf{43.23}\textcolor{red}{$\uparrow_{1.41}$}\\

   \midrule
   \multicolumn{7}{l}{\cellcolor{brown!20}\textit{\textbf{RL Fine-Tuned Constructor}}} \\
   \midrule
 
   Subgraph & 29.27 & 58.40 & 35.60 & 14.42 & 26.65 & 32.87 \\
   \cellcolor{teal!10}Subgraph + \texttt{DeepRefine-4B} &\underline{31.71} &\underline{62.14} &\underline{37.76} &\underline{16.94} &\textbf{30.57} &\underline{35.82}\textcolor{red}{$\uparrow_{2.95}$} \\
   \cellcolor{teal!20}Subgraph + \texttt{DeepRefine-8B} &\textbf{33.31} &\textbf{62.25} &\textbf{38.45} &\textbf{18.10} &\underline{29.44} &\textbf{36.31}\textcolor{red}{$\uparrow_{3.44}$} \\
   \cline{1-7}
  ToG & 29.88 & 60.34 & 48.56 & 19.06 & 23.64 &36.30  \\
   \cellcolor{teal!10}ToG + \texttt{DeepRefine-4B} &\underline{32.53} &\underline{63.51} &\textbf{54.83} &\underline{27.09}&\underline{29.05} &\underline{41.40}\textcolor{red}{$\uparrow_{5.10}$} \\
   \cellcolor{teal!20}ToG + \texttt{DeepRefine-8B} &\textbf{33.42} &\textbf{63.92} &\underline{54.28} &\textbf{27.82} &\textbf{29.45} &\textbf{41.78}\textcolor{red}{$\uparrow_{5.48}$} \\
   \cline{1-7}
   HippoRAG & \textbf{39.84}& 64.43 & 50.34 & 26.21 & 35.02 &43.17  \\
   \cellcolor{teal!10}HippoRAG + \texttt{DeepRefine-4B} &\underline{39.76}&\underline{65.66} & \textbf{53.97}&\underline{29.00}&\textbf{35.57}&\textbf{44.79}\textcolor{red}{$\uparrow_{1.62}$} \\
   \cellcolor{teal!20}HippoRAG + \texttt{DeepRefine-8B} & 39.73&\textbf{65.80}&\underline{53.41} &\textbf{29.40} &\underline{35.36}&\underline{44.74}\textcolor{red}{$\uparrow_{1.57}$}\\
   \cline{1-7}
   HippoRAG2 & 37.25 & 64.69 & 52.61 & 27.96 & 35.75 & 43.65 \\
   \cellcolor{teal!10}HippoRAG2 + \texttt{DeepRefine-4B} & \underline{38.81}&\underline{65.67}&\underline{57.29} & \textbf{29.34}&\textbf{36.99}&\textbf{45.62}\textcolor{red}{$\uparrow_{1.97}$} \\
   \cellcolor{teal!20}HippoRAG2 + \texttt{DeepRefine-8B} & \textbf{39.31}&\textbf{65.94}&\textbf{57.87} &\underline{28.16} &\underline{35.87}&\underline{45.43}\textcolor{red}{$\uparrow_{1.78}$} \\
 
   \midrule
   \multicolumn{7}{l}{\cellcolor{brown!20}\textit{\textbf{Graphify Constructor}}} \\
   \midrule
 
   Subgraph & 20.42 & 11.38 & 25.37 & \underline{11.69} & 11.02 & 15.98 \\
   \cellcolor{teal!10}Subgraph + \texttt{DeepRefine-4B} &\underline{21.71} &\underline{15.92}& \underline{25.45}&11.42 &\textbf{13.64} &\underline{17.63}\textcolor{red}{$\uparrow_{1.65}$} \\
   \cellcolor{teal!20}Subgraph + \texttt{DeepRefine-8B} &\textbf{22.31} &\textbf{17.35} &\textbf{28.64} &\textbf{12.25} & \underline{13.59}&\textbf{18.83}\textcolor{red}{$\uparrow_{2.85}$} \\
   \cline{1-7}
    ToG & 16.69 & 8.97 & 21.46 &11.73 &11.40 & 14.05 \\
   \cellcolor{teal!10}ToG + \texttt{DeepRefine-4B} &\underline{18.20}&\underline{10.91}&\underline{21.68} &\underline{12.59}&\underline{13.84} &\underline{15.44}\textcolor{red}{$\uparrow_{1.39}$} \\
   \cellcolor{teal!20}ToG + \texttt{DeepRefine-8B} &\textbf{20.27} &\textbf{15.11} &\textbf{29.51} &\textbf{14.56} & \textbf{13.86}&\textbf{18.66}\textcolor{red}{$\uparrow_{4.61}$} \\
   \cline{1-7}
    HippoRAG &13.23 &\underline{5.57}& 29.03&8.67 & 3.44& 11.99 \\
   \cellcolor{teal!10}HippoRAG + \texttt{DeepRefine-4B} &\underline{13.61} &\textbf{5.89} &\underline{30.18} &\textbf{9.25} &\underline{5.51}&\underline{12.89}\textcolor{red}{$\uparrow_{0.90}$} \\
   \cellcolor{teal!20}HippoRAG + \texttt{DeepRefine-8B} &\textbf{14.05} &5.26 &\textbf{36.54}& \underline{9.21}&\textbf{5.98}&\textbf{14.21}\textcolor{red}{$\uparrow_{2.22}$} \\
   \cline{1-7}
   HippoRAG2 & 13.62&6.19 & 27.98&8.10& 3.57& 11.89 \\
   \cellcolor{teal!10}HippoRAG2 + \texttt{DeepRefine-4B} &\underline{14.11} &\underline{6.56} &\underline{29.96} & \underline{9.99}&\underline{5.64}&\underline{13.25}\textcolor{red}{$\uparrow_{1.36}$}\\
   \cellcolor{teal!20}HippoRAG2 + \texttt{DeepRefine-8B} &\textbf{14.53} &\textbf{8.42} &\textbf{39.33} &\textbf{11.55} &\textbf{6.27}&\textbf{16.02}\textcolor{red}{$\uparrow_{4.13}$} \\
   \gbottomrule
   \end{tabular}}
   \label{tab:graph_retrievers_grouped}
   \end{table}
\vspace{-0.35cm}

\textcolor{opcolor}{\textbf{Insight \#1:}} \textbf{\textit{DeepRefine brings consistent downstream gains over various KB constructors and retrievers with better knowledge carrying and knowledge indexing capabilities.}} We refine the KBs constructed by the naive constructor, the RL fine-tuned constructor and the Graphify constructor. And we evaluate the performance of the original KBs and the refined ones with various retrievers. As shown in Table~\ref{tab:graph_retrievers_grouped}, with DeepRefine, the quality of the KBs can be stably improved for their better downstream performance in most cases. This suggests the refinement actions generated by the DeepRefine's policy can indeed improve the quality of KBs. And \texttt{DeepRefine-8B} can perform better than \texttt{DeepRefine-4B}, which demonstrates the effectiveness of the knowledge refinement can be improved by a larger scale model.
We have also demonstrate some cases about how DeepRefine can resolve issues in the constructed KBs in Appendix~\ref{sec:case_study}. Besides, we can also observe that DeepRefine introduces more performance gain for triple retrievers than text retrievers. This is because the refinement actions are triple-level operators and will not change the original text chunks. So the performance gains for triple retrievers are typically from both content and structure optimizations and the gains for text retrievers are only from the structure optimization to recall more accurate text chunks. This also demonstrates that DeepRefine can not only evolve the KB's knowledge carrying capabilities, but also the knowledge indexing quality. We also conduct held-out query evaluation in Appendix~\ref{appendix:heldout} to further validate that query-guided refinement accumulates as KB-level improvements rather than overfitting only to the queries used to drive the agent.

\begin{wrapfigure}{r}{0.35\textwidth}
  \centering
  \vspace{-0.25cm}
  \includegraphics[width=0.9\linewidth]{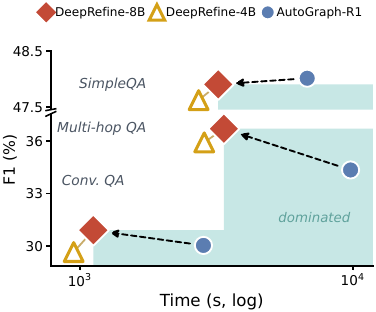}
  \vspace{-0.35cm}
  \caption{Effectiveness-efficiency comparisons of DeepRefine versus AutoGraph-R1.}
  \label{fig:compare}
  \vspace{-0.25cm}
  \end{wrapfigure}
\textcolor{opcolor}{\textbf{Insight \#2:}} \textbf{\textit{DeepRefine can introduce larger or competitive downstream gains than the reconstruction method.}} As shown in Table~\ref{tab:graph_retrievers_grouped} and Figure~\ref{fig:compare}, the downstream performance of the KBs constructed by a naive knowledge constructor can even outperform the fully reconstruction methods (AutoGraph-R1, which is also RL fine-tuned) in most cases after they are optimized by \texttt{DeepRefine-8B}. For example, in 2WikiQA, the performance of ``ToG+\texttt{DeepRefine-8B}'' with the naive constructor is much higher than that of ``ToG'' with the RL fine-tuned constructor (51.15 vs. 48.56). This is mainly because the downstream performance gain may not always be related to the whole KB content, but may only depend on some specific smaller regions of the KB. Regenerating the whole KB may introduce additional noise that may harm the downstream performance.

\begin{wraptable}{r}{0.52\textwidth}
  \centering
  \caption{Average time consumptions (s) \textdownarrow{} of DeepRefine vs.\ AutoGraph-R1.}
  \vspace{-0.15cm}
  \label{tab:efficiency_comparison}
  \resizebox{\linewidth}{!}{%
  \begin{tabular}{l|ccc}
  \gtoprule
  \theadrow
  \thcell{Methods} & \thcell{Simple QA} & \thcell{Multi-hop QA} & \thcell{Conversation QA} \\
  \gmidrule
  AutoGraph-R1 & 6,782.8 & 9,780.4 & 2,826.5 \\
  DeepRefine & \textbf{3,201.7} & \textbf{3,357.5} & \textbf{1,115.8} \\
  \gbottomrule
  \end{tabular}}
  \vspace{-0.15cm}
\end{wraptable}
\textcolor{opcolor}{\textbf{Insight \#3:}} \textbf{\textit{DeepRefine is more efficient than reconstructing the full KB.}} We collect the average time consumptions of both the refinement method DeepRefine and the reconstruction method AutoGraph-R1. As shown in Table~\ref{tab:efficiency_comparison}, on all three different kinds of benchmark data, DeepRefine exhibits obvious superiority with respect to the reconstruction method. The efficiency of DeepRefine comes from two aspects. First, instead of taking into account all documents and regenerating the whole KB, the agentic design of DeepRefine makes the model will only focus on the important and potentially problematic parts, which largely reduces the amount of information that is needed to be processed by the model. Further more, the design of generating refinement actions with the format of code can also further reduce the decoding time cost, which eliminates the need of inefficiently regenerating the whole subgraph content.
Consequently, according to Figure~\ref{fig:compare}, taking into account both efficiency and effectiveness, DeepRefine is a better choice to improve the quality of KBs than the reconstruction methods.

 \subsection{Ablation Study}

 To demonstrate the effectiveness and necessity of RL training for the DeepRefine, we first employ DeepRefine without RL training on the original benchmarks, which is denoted as \texttt{DeepRefine-8B/4B} \textit{w/o RL}. The performance is the average of two triple retrievers under the naive constructor, as triple retrievers are more consistent with the training settings and can reflect both of the knowledge indexing and knowledge carrying capabilities. We also further construct a benchmark and conduct experiments to verify the effectiveness of the RL training for resolving the incompleteness, incorrectness and redundancy issues, respectively. Benchmark construction details are elaborated in Appendix~\ref{sec:additional_experiments}.

 \begin{table}[h]
  \centering
  \caption{Performance comparisons between GBD reward finetuned DeepRefine and the DeepRefine without RL training under the naive constructor.}
  \resizebox{0.8\textwidth}{!}{%
  \begin{tabular}{l|ccccc|c}
  \gtoprule
  \theadrow
  & \multicolumn{2}{c}{\thcell{Simple QA}} & \multicolumn{2}{c}{\thcell{Multi-hop QA}} & \multicolumn{1}{c}{\thcell{Conversation}} & \multicolumn{1}{c}{\thcell{Overall}} \\
  \theadrow
  \multirow{-2}{*}{\thcell{Variants}} & \thcell{NQ*} & \thcell{PopQA*} & \thcell{2WikiQA} & \thcell{Musique} & \thcell{LOCOMO} & \thcell{Avg.} \\
  \gmidrule
  \cellcolor{teal!10}\texttt{DeepRefine-4B} & \textbf{29.38} & \textbf{59.68} & \textbf{42.60} & \textbf{20.83} & \textbf{27.09} & \textbf{35.92} \\
  \texttt{DeepRefine-4B} \textit{w/o RL} & 28.74 & 58.89 & 41.37 &19.86 & 26.89 &35.15 \\
  \hline
  \cellcolor{teal!20}\texttt{DeepRefine-8B} &\textbf{30.29} & \textbf{59.93} &\textbf{44.77}  &\textbf{21.71}  & \textbf{27.02} & \textbf{36.74}\\
  \texttt{DeepRefine-8B} \textit{w/o RL} & 29.19 & 59.73 &42.71 & 20.62 & 25.08 &35.47 \\
  \gbottomrule
  \end{tabular}}
  \label{tab:ablation_8B}
  \end{table}

\begin{wrapfigure}{r}{0.55\textwidth}
  \centering
  \includegraphics[width=\linewidth]{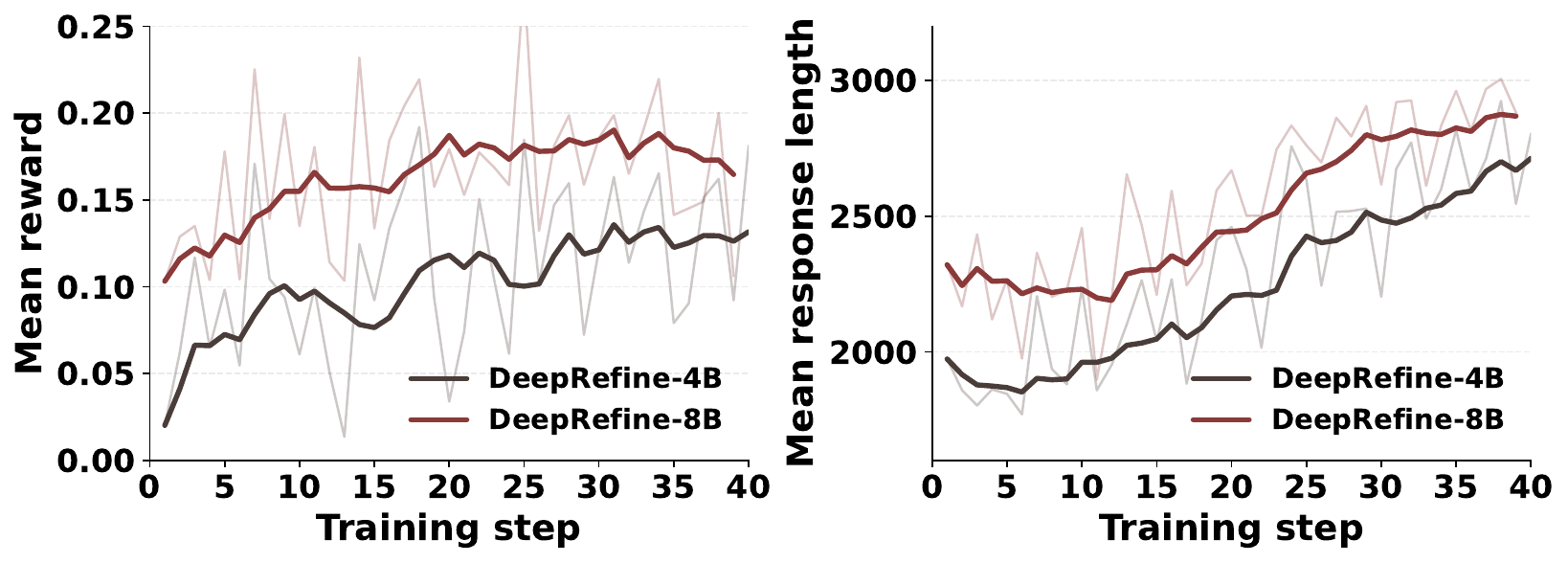}
  \caption{GBD reward and response length curves of \texttt{DeepRefine-4B/8B} during RL training.}
  \label{fig:rl_training_rewards}
  \end{wrapfigure}
  
\textcolor{opcolor}{\textbf{Insight \#4:}} \textbf{\textit{RL training with GBD reward can introduce consistent performance gain.}} As shown in Table~\ref{tab:ablation_8B}, removing RL training would lead to lower performance, especially for multi-hop QA and conversation QA. This is mainly because without RL optimization, it would be hard for the agent to fully utilize the interaction history in the knowledge refinement task. Besides, the refinement policy without optimization would also be not very useful for the downstream tasks. As shown in Figure~\ref{fig:rl_training_rewards}, we also demonstrate the training dynamics with average GBD reward and response length curves of both two different size models. As the training progress goes on, the reward gains are steadily increasing. And the model will also first learn to reduce the response length by judging with more answerable queries to avoid introduce reward deduction, which is a kind of reward hacking, but then it learns to explore longer reasoning trajectories and reach a better refinement policy $\pi_{\theta}$.

\begin{wrapfigure}{r}{0.55\textwidth}
  \centering
  \includegraphics[width=\linewidth]{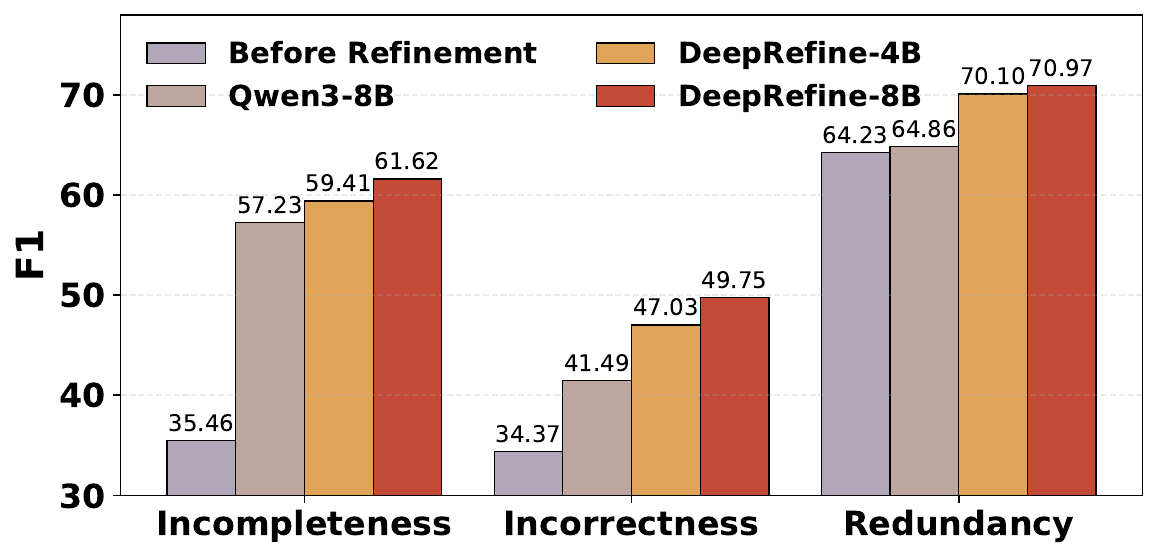}
  \caption{Effectiveness of DeepRefine and base models on resolving three types of issues with the subgraph retriever.}
  \label{fig:refinement_effectiveness}
  \vspace{-0.25cm}
  \end{wrapfigure}

\textcolor{opcolor}{\textbf{Insight \#5:}} \textbf{\textit{The base model is preliminary effective for resolving incompleteness and incorrectness, but to a lesser degree for resolving redundancy issues.}} As shown in Figure~\ref{fig:refinement_effectiveness}, the base model \texttt{Qwen3-8B} that employs the DeepRefine's agentic framework can resolve some incompleteness and incorrectness issues to improve the performance to some degree. But it can hardly resolve the redundancy issues. Specifically, the incompleteness issue degrades performance very hard and can also be improved the most easily.
The incorrectness issue can hurt performance the most, and it is harder to be resolved. Because it requires the model to have the capability to identify the incorrect information and then correct it according to the source passages. Besides, the redundancy issue degrades the downstream performance the mildest, and it is also the hardest to be resolved. Because it requires the model to do the coreference resolution and disambiguation, which is more challenging than the other two tasks.

\textcolor{opcolor}{\textbf{Insight \#6:}} \textbf{\textit{RL training with GBD reward can introduce better effectivenss for resolving incompleteness, incorrectness and redundancy issues.}} As shown in Figure~\ref{fig:refinement_effectiveness}, both of the \texttt{DeepRefine-8B} and \texttt{DeepRefine-4B} that are optimized on the GBD reward is more effective than the base model \texttt{Qwen3-8B} in resolving the incompleteness, incorrectness and redundancy issues. Specifically, the superiority of \texttt{DeepRefine-8B} and \texttt{DeepRefine-4B} over \texttt{Qwen3-8B} is more significant for the incorrectness and redundancy issues, which are harder to be resolved as we mentioned before. The base model has almost no effectiveness for fixing the redundancy issue (only $\sim 0.63\%$), however, the RL fine-tuned models can improve the performance to a certain extent ($\sim 6\%$), which is $\text{x}10$ better than the base model. And the RL fine-tuned models also exhibit superior performance over the base model for resolving the incompleteness issues. This is because the GBD reward can provide an accurate and informative feedback for the agentic refinement process during the RL training, which helps the model to learn a both effective and general refinement policy for resolving various kinds of issues. And the results can also suggest that not only our agentic knowledge refinement framework can localize the problematic parts in KBs, but also the RL fine-tuning strategy can further improve the refinement effectiveness through solving the three potential issues.
 \section{Conclusion}
 In this paper, we propose DeepRefine, an RL framework for agentic knowledge refinement. DeepRefine is designed to refine pre-constructed structured KBs by leveraging the multiple interactions with the KB, which is trained with a newly designed GBD reward finetuned RL policy. DeepRefine makes the KBs evolvable with historical user interactions. Experimental results show that DeepRefine can consistently improve the quality of the constructed KB and achieve better downstream performance compared with that of the original KB.

\bibliography{colm2026_conference}
\bibliographystyle{colm2026_conference}

\clearpage
\appendix
\begin{center}
  \LARGE \bf {Appendix of \textbf{DeepRefine}}
\end{center}
\etocdepthtag.toc{mtappendix}
\etocsettagdepth{mtchapter}{none}
\etocsettagdepth{mtappendix}{subsection}
\tableofcontents
\clearpage

\section{Coverage-Based Query Selection}
\label{sec:coverage_based_query_selection}
\par
\begin{wraptable}{r}{0.3\textwidth}
  \centering
  \caption{Hyperparameters for greedy maximum-coverage query selection.}
  \label{tab:coverage_hyperparams}
  \begin{tabular}{lc}
  \gtoprule
  \theadrow
  \thcell{Parameter} & \thcell{Value} \\
  \gmidrule
  $k$ & $10$ \\
  $m$ & $100$ / $500$ \\
  $B$ & $1000$ \\
  $\rho$ & $1.0$ / $0.8$ \\
  \gbottomrule
  \end{tabular}
\end{wraptable}
Considering there could be overlappings among the effective regions of the refinement processes with different queries in each benchmark dataset, to make the knowledge refinement process both efficient and effective, we select a small subset of queries before refinement using a \textbf{greedy maximum-coverage procedure} over query-related triples. For each candidate query respect to the full knowledge base in a benchmark dataset, we obtain a set of triples by (i) taking the top-$k$ related triples with their sentence embeddings, and (ii) augmenting with up to $m$ additional triples drawn from the one-hop neighborhoods of the entities appearing in those triples, where neighborhood triples are ranked by embedding similarity to the query and the highest-scoring ones are kept. Treating each distinct triple as a coverage element, we greedily choose queries that maximize the number of newly covered elements until either a budget of $B$ queries is reached or the fraction of covered elements in the union of all candidates' sets exceeds a target threshold $\rho$. Only the selected queries are then passed through the refinement pipeline to update the shared knowledge base. Then all benchmark queries are evaluated on the resulting knowledge base with the standard RAG pipeline. The corresponding hyperparameter values are listed in Table~\ref{tab:coverage_hyperparams}. For LOCOMO dataset, $m$ is set to $100$ and $\rho$ is set to $1.0$. For other larger datasets, $m$ is set to $500$ and $\rho$ is set to $0.8$.

\section{Training and Inference Algorithms}
\label{sec:training_inference_algorithms}

We summarize the three-step reasoning rollout, GRPO training with the GBD reward, and deployment-time inference in Algorithms~\ref{alg:reasoning}--\ref{alg:inference}. As shown in Algorithm~\ref{alg:reasoning}, note that according to prior work~\citep{jin2025search}, 

\begin{algorithm}[H]
\caption{$\textsc{DeepRefineRollout}\!\left(q,\,\mathcal{G}_{\text{f}},\,\pi_{\theta},\,L_{\max},\,N,\,M\right)$}
\label{alg:reasoning}
\begin{algorithmic}[1]
\Require Query $q$, per-query draft KB $\mathcal{G}_{\text{f}}$ (each query has its own separated sub-KB), policy $\pi_{\theta}$,
         hop limit $L_{\max}$, top-$N$ / top-$M$ retrieval sizes.
\Ensure Multi-turn reasoning output $\mathbf{o}$ with refinement actions $\mathcal{A}_{\text{q}}$.

\State $\mathcal{G}^{(0)}_{\text{q}} \leftarrow \textsc{Top-}k(q, \mathcal{G}_{\text{f}}, N)$;\;
       $\mathcal{H}_{\text{q}} \leftarrow \emptyset$;\;
       $i \leftarrow 0$
\While{$i < L_{\max}$}
    \State $\mathcal{J}^{(i)}_{\text{q}}, \mathbf{o} \leftarrow \textsc{Judge}_{\theta}(q, \mathcal{G}^{(i)}_{\text{q}})$
    \Comment{output in \textcolor{opcolor}{\texttt{<judge>}} tags}
    \State Append $(q, \mathcal{G}^{(i)}_{\text{q}}, \mathcal{J}^{(i)}_{\text{q}})$ to $\mathcal{H}_{\text{q}}$
    \If{$\mathcal{J}^{(i)}_{\text{q}} = \text{Yes}$}
        \Break
    \EndIf
    \State $\mathcal{G}^{(i+1)}_{\text{q}} \leftarrow \mathcal{G}^{(i)}_{\text{q}} \cup
           \textsc{Top-}k\!\big(q, \mathcal{G}_{\text{cand}}^{(i+1)}, M\big)$
    \Comment{one-hop neighbor collection in Eqs.~\eqref{eq:collect_neighbors}--\eqref{eq:merge_subgraph}}
    \State $i \leftarrow i + 1$
\EndWhile
\If{$\mathcal{J}^{(0)}_{\text{q}} \neq \text{Yes}$}
    \State $\mathcal{I}_{\text{q}}, \mathbf{o} \leftarrow \textsc{Abduce}_{\theta}(\mathcal{H}_{\text{q}})$
    \Comment{output in \textcolor{opcolor}{\texttt{<abduction>}} tags}
    \State $\mathcal{A}_{\text{q}}, \mathbf{o} \leftarrow \textsc{Refine}_{\theta}(\cdot \mid \mathcal{G}^{(i)}_{\text{q}}, \mathcal{I}_{\text{q}})$
    \Comment{output in \textcolor{opcolor}{\texttt{<refinement>}} tags}
\Else
    \State $\mathcal{A}_{\text{q}} \leftarrow \emptyset$
\EndIf

\end{algorithmic}
\end{algorithm}

\begin{algorithm}[H]
\caption{$\textsc{DeepRefineGRPO}\!\left(\mathcal{Q}_{\text{train}},\,\{\mathcal{G}_{\text{f}}^{(q)}\},\,\pi_{\theta},\,G,\,\epsilon\right)$}
\label{alg:training}
\begin{algorithmic}[1]
\Require Training queries $\mathcal{Q}_{\text{train}}$, per-query draft KBs $\{\mathcal{G}_{\text{f}}^{(q)}\}$,
         policy $\pi_{\theta}$, frozen reader model $\textsc{Reader}(\cdot)$, group size $G$, clipping hyperparameter $\epsilon$.
\Ensure Optimized refinement policy $\pi_{\theta}$.

\For{each GRPO training step}
    \For{each query $q$ in the mini-batch sampled from $\mathcal{Q}_{\text{train}}$}
        \State $\mathcal{G}_{\text{f}} \leftarrow \mathcal{G}_{\text{f}}^{(q)}$;\;
               $A \leftarrow$ golden answer of $q$;\; $A_{\text{draft}} \leftarrow \textsc{Reader}(q, \mathcal{G}_{\text{f}})$
        \For{$i = 1$ \textbf{to} $G$}
            \State $\mathbf{o}_i, \mathcal{A}_{\text{q}}^{(i)} \leftarrow
                   \textsc{DeepRefineRollout}(q, \mathcal{G}_{\text{f}}, \pi_{\theta_{\text{old}}})$
            \Comment{Alg.~\ref{alg:reasoning}}
            \State $\hat{\mathcal{G}}^{(i)}_{\text{f}} \leftarrow
                   \textsc{ApplyActions}(\mathcal{G}_{\text{f}}, \mathcal{A}_{\text{q}}^{(i)})$
            \State $A^{(i)}_{\text{refined}} \leftarrow \textsc{Reader}(q, \hat{\mathcal{G}}^{(i)}_{\text{f}})$
            \State $R_i \leftarrow \text{F1}(A^{(i)}_{\text{refined}}, A) - \text{F1}(A_{\text{draft}}, A)$
        \EndFor
        \State $\mu_R \leftarrow \frac{1}{G}\sum_{i=1}^{G} R_i$;\;
               $\sigma_R \leftarrow \textsc{Std}(\{R_i\}_{i=1}^{G})$
        \For{$i = 1$ \textbf{to} $G$}
            \For{$t = 1$ \textbf{to} $|\mathbf{o}_i|$}
                \State $\hat{A}_{i,t} \leftarrow (R_i - \mu_R)/\sigma_R$
                \State $r_{i,t}(\theta) \leftarrow
                       \frac{\pi_{\theta}(o_{i,t} \mid q, \mathbf{o}_{i,<t})}
                            {\pi_{\theta_{\text{old}}}(o_{i,t} \mid q, \mathbf{o}_{i,<t})}$
                \State Accumulate clipped GRPO objective using $r_{i,t}(\theta)$ and $\hat{A}_{i,t}$
            \EndFor
        \EndFor
    \EndFor
    \State Update $\theta$ by gradient ascent on the GRPO objective
\EndFor
\end{algorithmic}
\end{algorithm}

\begin{algorithm}[H]
\caption{$\textsc{DeepRefineInfer}\!\left(\mathcal{G}_{\text{f}},\,\mathcal{Q},\,\pi_{\theta},\,k,\,m,\,B,\,\rho\right)$}
\label{alg:inference}
\begin{algorithmic}[1]
\Require Pre-constructed KB $\mathcal{G}_{\text{f}}$, query set $\mathcal{Q}$, frozen policy $\pi_{\theta}$,
         coverage hyperparameters $(k, m, B, \rho)$.
\Ensure Evolved KB $\mathcal{G}$ for downstream RAG.

\State $\mathcal{G} \leftarrow \mathcal{G}_{\text{f}}$
\State $\mathcal{Q}_{\text{refine}} \leftarrow
       \textsc{GreedyCoverageSelect}(\mathcal{Q}, \mathcal{G}, k, m, B, \rho)$
\Comment{Appendix~\ref{sec:coverage_based_query_selection}}
\For{each $q \in \mathcal{Q}_{\text{refine}}$}
    \State $\mathbf{o}, \mathcal{A}_{\text{q}} \leftarrow
           \textsc{DeepRefineRollout}(q, \mathcal{G}, \pi_{\theta})$
    \Comment{Alg.~\ref{alg:reasoning}}
    \If{$\mathcal{A}_{\text{q}} \neq \emptyset$}
        \State $\mathcal{G} \leftarrow \textsc{ApplyActions}(\mathcal{G}, \mathcal{A}_{\text{q}})$
    \EndIf
\EndFor
\end{algorithmic}
\end{algorithm}

\section{Benchmark Construction}
\label{sec:additional_experiments}

To further evaluate the effectiveness of DeepRefine on resolving incompleteness, incorrectness or redundancy issues, we provide a rigorous evaluation. We construct a benchmark characterized by explicit error types and diverse corruption patterns. By editing the knowledge bases with controlled defects that mirror real-world failure modes, it enables a precise assessment of a model's performance in defect localization and refinement processes across various error types.

We constuct the benchmark based on the HotpotQA (Dev-Distractor)~\citep{yang2018hotpotqa} dataset, which is different from the training set of HotpotQA that is used for our RL training. It consists of 501 samples, with 167 samples dedicated to each of the three systematic error types defined in our study. Each data point includes the original query, the corrupted knowledge base, and comprehensive metadata (e.g., corruption details and action counts). The construction pipeline utilizes \texttt{Qwen2.5-32B-Instruct} (denoted as the corruption model) for generating defects, \texttt{Qwen2.5-3B-Instruct} as the naive knowledge base constructor, and \texttt{Qwen3-Embedding-0.6B} for vector-based retrieval. \texttt{Qwen2.5-7B-Instruct} serves as the reader model for RAG-based evaluation.

\textit{\textbf{Error Types and Corruption Strategies.}} For each error type, we apply a specific corruption strategy to a high-quality knowledge base: \textit{Incompleteness}: We identify and remove critical triples that are essential for the reasoning path, creating "missing evidence" scenarios. \textit{Incorrectness}: We modify correct knowledge items into incorrect claims based on the original context, simulating low-confidence or imprecise information. \textit{Redundancy}: This is achieved via two methods: (1) Inserting semantically similar but redundant triples to introduce ambiguity. (2) Replacing entities in existing triples with aliases or synonymous names to create coreference resolution challenges.

\textit{\textbf{Construction Pipeline.}} The construction pipeline follows a rigorous multi-stage pipeline. Initially, for each sample, we construct an individual KB based on its context. We then perform RAG. Only samples achieving a F1 score higher than $0.95$ are retained. This ensures that the initial KB is relatively accurate and contains sufficient information to answer the query. Then, for the filtered samples, the corruption process is specifically targeted at the top-$N$ triples retrieved during the previous RAG stage. This ensures that the induced defects are concentrated on the critical knowledge components most relevant to the query, rather than introducing peripheral noise. Within this focused scope, the corruption model generates a series of actions. To ensure the generation of ideal, high-quality defects, we designed specialized prompt templates and carefully curated few-shot samples tailored to each specific error type. These actions are subsequently parsed into three predefined operators: \textcolor{opcolor}{\texttt{insert\_edge()}}, \textcolor{opcolor}{\texttt{delete\_edge()}}, and \textcolor{opcolor}{\texttt{replace\_node()}}. The action space is designed to be symmetric with DeepRefine's refinement action set to facilitate direct performance mapping. Notably, the replace\_node() operator is restricted to acting upon a single, specific edge. This fine-grained precision allows us to create nuanced coreference resolution issues and subtle semantic defects that test the model's diagnostic depth. To maintain the structural integrity and ensure the refinability of the knowledge base, the total number of corruption actions is strictly limited to 5 per sample. Then we perform a second RAG evaluation on the corrupted KB. Only samples where the RAG F1 score drops below 0.6 are included in the final benchmark. This ensures that the corruption is effective, which suggests that it significantly impairs the model's ability to answer the query without refinement.

\section{Held-out Query Evaluation}~\label{appendix:heldout}
\begin{wraptable}{r}{0.45\textwidth}
\centering
\small
\caption{Query-select refinement on 2WikiMultihopQA and Musique, evaluated on queries excluded from the refine subset.}
\label{tab:2wiki-select-test}
\resizebox{\linewidth}{!}{%
\begin{tabular}{l|cc}
\toprule
Methods &2WikiQA&Musique\\
\midrule
Subgraph & 32.51 &  13.51\\
\cellcolor{teal!10}Subgraph + DeepRefine-4B& \underline{34.38}\textcolor{red}{$\uparrow_{1.87}$} & \textbf{20.08}\textcolor{red}{$\uparrow_{6.57}$}\\
\cellcolor{teal!20}Subgraph + DeepRefine-8B& \textbf{34.53}\textcolor{red}{$\uparrow_{2.02}$} & \underline{18.07}\textcolor{red}{$\uparrow_{4.56}$}\\
\midrule
ToG & 42.58 & 16.78  \\
\cellcolor{teal!10}ToG + DeepRefine-4B & \textbf{44.85}\textcolor{red}{$\uparrow_{2.27}$} & \textbf{29.71}\textcolor{red}{$\uparrow_{12.93}$} \\
\cellcolor{teal!20}ToG + DeepRefine-8B & \underline{44.11}\textcolor{red}{$\uparrow_{1.53}$} & \underline{22.64}\textcolor{red}{$\uparrow_{5.86}$} \\
\bottomrule
\end{tabular}
}
\end{wraptable}
In our experiments in Section~\ref{sec:experimental_settings}, the results have suggested that refining the KBs with only a small set of queries (from $\sim 10\%$ to $\sim 40\%$ of the queries in the benchmarks) can lead to a general performance gain under all queries in the benchmarks. In this section, to further demonstrate the refinement actions applied to the KBs are also useful for different query sets, we refine only on the selected subset, then evaluate ToG and Subgraph retrievers on the complementary holdout, comparing the original KB with the KB after refinement.

We use the same coverage-based selector as in the main experiments, with
$(\rho,B)=(0.8,500)$ on 2WikiMultihopQA and MuSiQue (1{,}000 queries each).
This selects 446  queries for \texttt{DeepRefine-4B / 8B} on 2Wiki (holdout
554) and 413 on MuSiQue (holdout 587). As shown in Table~\ref{tab:2wiki-select-test}, the same refinement actions remain useful for queries that were never selected. Both retrievers improve on both datasets under \texttt{DeepRefine-4B / 8B}. The holdout result therefore supports that query-guided refinement accumulates as KB-level improvements rather than overfitting only to the queries used to drive the agent.

\section{Case Study}
\label{sec:case_study}
In this section, we will provide a case study to demonstrate the effectiveness of DeepRefine. We will give some cases to show how DeepRefine can refine the knowledge base to deal with the potential incompleteness, incorrectness or redundancy issues.

\subsection{Case 1: Incompleteness}\label{sec:case_incompleteness}
Incompleteness is the most common issue in the knowledge base. It is not possible to connect all potentially relevant knowledge items in the real world, which makes it necessary to do the completions that are useful for the downstream tasks.
As shown in Table~\ref{tab:case_incompleteness}, the question asks for the 2010 population of the city that is popular with tourists. In the original query-related knowledge items, there are tourism-related clues and several population facts, but the key bridge from the target city to its 2010 population is missing. DeepRefine addresses this incompleteness by inserting series of missing population edges (e.g., city $\rightarrow$ ``population in 2010'' $\rightarrow$ value), which shortens the reasoning path and makes the answer directly retrievable.
  
\begin{table}[H]
  \centering
  \begin{tabular}{|>{\raggedright\arraybackslash}p{0.35\textwidth}|>{\raggedright\arraybackslash}p{0.15\textwidth}|>{\raggedright\arraybackslash}p{0.4\textwidth}|}
  \hline
  \theadrow
  \textit{\textbf{Original Passages}} & \textit{\textbf{Query}} & \textit{\textbf{Query-Related Triples}} \\
  \hline
  ...\newline
  Instead, Nanjing, as a popular tourist destination, hosts a series of government-organised events throughout the year... \newline
  The total population of the City of Nanjing reached 8.005 million in 2010. The statistics in 2011 estimated the total population to be 8.11 million...\newline
  Camdenton is a city in Camden County, Missouri, United States. The population was 3,718 at the 2010 census...\newline
  ... & ~\newline What is the population in 2010 of the city popular with tourists?&...\newline$<$\textit{`Water tourism inside Strasbourg',`attracts',`hundreds of thousands of tourists yearly'}$>$,\newline$<$\textit{`Oklahoma',`included four cities over 100,000 in population',`in 2010'}$>$,\newline$<$\textit{`New York City',`population reached all-time high',`in the 2010 Census'}$>$,\newline$<$\textit{`Nanjing',`has annual attraction',`thousands of tourists'}$>$,\newline$<$\textit{`Oklahoma City',`had the largest metropolitan area in the state in 2010',`with 1,252,987 people'}$>$\\
  \hline
  \theadrow
  \multicolumn{3}{|l|}{\cellcolor{Gray}\textit{\textbf{Refinement Actions}}} \\
  \hline
  \multicolumn{3}{|p{0.9\textwidth}|}{\textcolor{opcolor}{\texttt{<refinement>}}insert\_edge(\textit{`Palermo'}, \textit{`population in 2010'}, \textit{`1.2 million'})$|$insert\_edge(\textit{`Nanjing'}, \textit{`population in 2010'}, \textit{`8.005 million'})$|$insert\_edge(\textit{`Tranquillity'}, \textit{`population in 2010'}, \textit{`799'})$|$insert\_edge(\textit{`Quincy'}, \textit{`population in 2010'}, \textit{`7,972'})$|$insert\_edge(\textit{`York'}, \textit{`population in 2010'}, \textit{`910'})$|$insert\_edge(\textit{`Oklahoma City'}, \textit{`population in 2010'}, \textit{`620,602'})$|$insert\_edge(\textit{`New York City'}, \textit{`population in 2010'}, \textit{`8.17 million'})$|$insert\_edge(\textit{`Seattle'}, \textit{`population in 2010'}, \textit{`608,000'})$|$insert\_edge(\textit{`Oklahoma'}, \textit{`population in 2010'}, \textit{`3.8 million'})$|$\textcolor{opcolor}{\texttt{</refinement>}}} \\
  \hline
  \end{tabular}
  \caption{The case of refinement for incompleteness issue.}
  \label{tab:case_incompleteness}
\end{table}
\vspace{-0.5cm}

\subsection{Case 2: Incorrectness}\label{sec:case_incorrectness}
Errors usually come from the incorrect or conflicting information in the original passages.
As shown in Table~\ref{tab:case_errors}, the error in this case is a factual contradiction introduced by a misalignment between the knowledge base content and the ground truth evidence. Specifically, the retrieved query-related triples contains a wrong edge $<$\textit{`Kree Harrison', `took runner-up spot', `American Idol (season 3)'}$>$ which contradicts the gold evidence stating that Diana DeGarmo was the runner-up of that season. If this incorrect edge is preserved, the model will inevitably reach the wrong answer regarding the contestant's identity and subsequent reasoning hops.

DeepRefine handles this through its refinement mechanism: it identifies the mismatch and applies \textcolor{opcolor}{\texttt{delete\_edge}} as well as \textcolor{opcolor}{\texttt{insert\_edge}} to restore the correct factual link $<$\textit{`Diana DeGarmo', `runner-up of', `American Idol (season 3)'}$>$ while effectively overriding the incorrect noise. This ensures the reasoning chain remains faithful to the supporting paragraphs.

\begin{table}[H]
  \centering
  \begin{tabular}{|>{\raggedright\arraybackslash}p{0.3\textwidth}|>{\raggedright\arraybackslash}p{0.2\textwidth}|>{\raggedright\arraybackslash}p{0.4\textwidth}|}
  \hline
  \theadrow
  \textit{\textbf{Original Passages}} & \textit{\textbf{Query}} & \textit{\textbf{Query-Related Triples}} \\
  \hline
  ...\newline
  Fantasia and \textbf{Diana DeGarmo} were the last two finalists, and Fantasia was crowned as the winner. The third season was won by Fantasia Barrino, who defeated \textbf{Diana DeGarmo}...
  ...\newline
  & ~\newline Which season was the runner up of the third season of American Idol on?
  & ... \newline $<$\textit{`American Idol (season 3)', `winner', `Fantasia Barrino'}$>$
    , \newline $<$\textit{`American Idol (season 3)', `runner-up', `Kree Harrison'}$>$
    , \newline $<$\textit{`Kree Harrison', `took runner-up spot', `American Idol'}$>$,\newline... \\
  \hline
  \theadrow
  \multicolumn{3}{|l|}{\cellcolor{Gray}\textit{\textbf{Refinement Actions}}} \\
  \hline
  \multicolumn{3}{|p{0.9\textwidth}|}{\textcolor{opcolor}{\texttt{<refinement>}}delete\_edge(\textit{`American Idol (season 3)'}, \textit{`runner-up'}, \textit{`Kree Harrison'}) $|$ insert\_edge(\textit{`American Idol (season 3)'}, \textit{`runner-up'}, \textit{`Diana DeGarmo'}) $|$ insert\_edge(\textit{`Diana DeGarmo'}, \textit{`runner-up of'}, \textit{`American Idol (season 3)'})\textcolor{opcolor}{\texttt{</refinement>}}} \\
  \hline
  \end{tabular}
  \caption{The case of refinement for Incorrectness issue involving factual contradiction.}
  \label{tab:case_errors}
\end{table}
\vspace{-0.9cm}

\subsection{Case 3: Redundancy}\label{sec:case_incorrectness}
We mainly focus on two types of redundancy issues: coreference resolution and disambiguation. And we will give some examples to show how DeepRefine can deal with these issues.

\textbf{\textit{Coreference resolution.}} As shown in Table~\ref{tab:case_redundancy}, the original knowledge base contains a coreference issue: According to the original passages, \textit{`the girl's phone number'} is extracted as an knowledge item in the query-related triples, and is referenced by both James. However, as the query is about whose phone number did James receive during the beach outing, even with some contextual information in the query-related triples, the models still cannot figure out who is \textit{`the girl'} mentioned in the triple. DeepRefine can correctly resolve this issue by first locating the potentially problematic region, then replacing \textit{`the girl's phone number'} with \textit{`Samantha's phone number'} and inserting a new edge between James and Samantha's phone number.

\begin{table}[H]
  \centering
  \begin{tabular}{|>{\raggedright\arraybackslash}p{0.4\textwidth}|>{\raggedright\arraybackslash}p{0.15\textwidth}|>{\raggedright\arraybackslash}p{0.35\textwidth}|}
  \hline
  \theadrow
  \textit{\textbf{Original Passages}} & \textit{\textbf{Query}} & \textit{\textbf{Query-Related Triples}} \\
  \hline
  ...\newline
  {\bf James: }Yesterday I took my three dogs to a beach outing to have fun and bond with other dogkeepers.\newline
  {\bf John: }Cool! Surely you gained a new experience from communicating with other dog lovers!\newline
  {\bf James: }Yes, we had fun and I even met one beautiful girl. I'm thinking of asking her out on a date! She left me her phone number, I think I'll call tomorrow.\newline
  {\bf John: }Wow! That's cool, what's her name? Be sure to call her, everything will work out!\newline
  {\bf James: }She is Samantha. I'll definitely call her!\newline
  {\bf John: }Yoohoo! Hope you have a wonderful time!\newline
  ... & ~\newline Whose phone number did James receive during the beach outing?&...\newline$<$\textit{`James',`left',`the girl's phone number'}$>$,\newline$<$\textit{`James',`announced',`decision to move in together with Samantha'}$>$, \newline$<$\textit{`James',`took his three dogs to',`beach outing'}$>$, \newline$<$\textit{`James',`received',`congrats James! That's awesome.'}$>$, \newline$<$\textit{`John',`wishing',`James a great time'}$>$,\newline...\\
  \hline
  \theadrow
  \multicolumn{3}{|l|}{\cellcolor{Gray}\textit{\textbf{Refinement Actions}}} \\
  \hline
  \multicolumn{3}{|p{0.9\textwidth}|}{\textcolor{opcolor}{\texttt{<refinement>}}replace\_node(\textit{`the girl's phone number'}, \textit{`Samantha's phone number'})$|$insert\_edge(\textit{`James'}, \textit{`received'}, \textit{`Samantha's phone number'})$|$\textcolor{opcolor}{\texttt{</refinement>}}} \\
  \hline
  \end{tabular}
  \caption{The case of refinement for coreference resolution issue.}
  \label{tab:case_redundancy}
\end{table}

\textbf{\textit{Disambiguation.}} As shown in Table~\ref{tab:case_disambiguation}, the original knowledge base contains a potential disambiguation issue: The person in the original document, \textit{`Ray Taylor'}, is extracted exactly as the same name string in the knowledge base. However, this simple extraction strategy may cause the ambiguity of the knowledge base, for example, there could be another person with the same name string in the knowledge base, which may introduce some structural noise to the related triples. DeepRefine can correctly resolve this issue by first locating the potentially problematic subgraph, then replacing \textit{`Ray Taylor'} with \textit{`Ray Taylor (1888-12-01 to 1952-02-15)'}, which smartly handles the disambiguation issue by adding more specific information to the knowledge items in the query-related triples. In this way, if any new people with the same name appear, the knowledge base will not confuse them.

\begin{table}[H]
  \centering
  \begin{tabular}{|>{\raggedright\arraybackslash}p{0.35\textwidth}|>{\raggedright\arraybackslash}p{0.15\textwidth}|>{\raggedright\arraybackslash}p{0.40\textwidth}|}
  \hline
  \theadrow
  \textit{\textbf{Original Passages}} & \textit{\textbf{Query}} & \textit{\textbf{Query-Related Triples}} \\
  \hline
  ...\newline
  Ray Taylor (1 December 1888 - 15 February 1952) was an American film director. He directed 159 films between 1926 and 1949. His debut was the 1926 film serial \"Fighting with Buffalo Bill\". Ravina is an Indian actress who acted in Dhallywood movies. She acted in\" Praner Cheye Priyo\" which film is considered as turning point of the career of Riaz.\newline
  ... & ~\newline Which film has the director died later, Modern Husbands or The Fighting Vigilantes?&...\newline$<$\textit{`"Fighting with Buffalo Bill"',`was directed by',`Ray Taylor'}$>$,\newline$<$\textit{`Modern Husbands',`starring',`Olinda Bozán'}$>$,\newline$<$\textit{`Ray Taylor',`debut film',`"Fighting with Buffalo Bill"'}$>$,\newline$<$\textit{`Ray Taylor',`directed',`159 films'}$>$,\newline$<$\textit{`Ray Taylor',`lived from',`1888-12-01 to 1952-02-15'}$>$,\newline$<$\textit{`Ray Taylor',`was a',`American film director'}$>$\newline...\\
  \hline
  \theadrow
  \multicolumn{3}{|l|}{\cellcolor{Gray}\textit{\textbf{Refinement Actions}}} \\
  \hline
  \multicolumn{3}{|p{0.95\textwidth}|}{\textcolor{opcolor}{\texttt{<refinement>}}insert\_edge(\textit{`Modern Husbands'}, \textit{`directed-by'}, \textit{`Luis Bayón Herrera'})$|$insert\_edge(\textit{`The Fighting Vigilantes'}, \textit{`directed-by'}, \textit{`Ray Taylor'})$|$insert\_edge(\textit{`Modern Husbands'}, \textit{`released-on'}, \textit{`1948-01-01'})$|$insert\_edge(\textit{`The Fighting Vigilantes'}, \textit{`released-on'}, \textit{`1947-11-15'})$|$replace\_node(\textit{`Ray Taylor'}, \textit{`Ray Taylor (1888-12-01 to 1952-02-15)'})$|$\textcolor{opcolor}{\texttt{</refinement>}}} \\
  \hline
  \end{tabular}
  \caption{The case of refinement for disambiguation issue.}
  \label{tab:case_disambiguation}
\end{table}

\section{Rationality of the Action Space Design}\label{appendix:action_space}

\begin{figure}[H]
   \centering
   \includegraphics[width=0.9\linewidth]{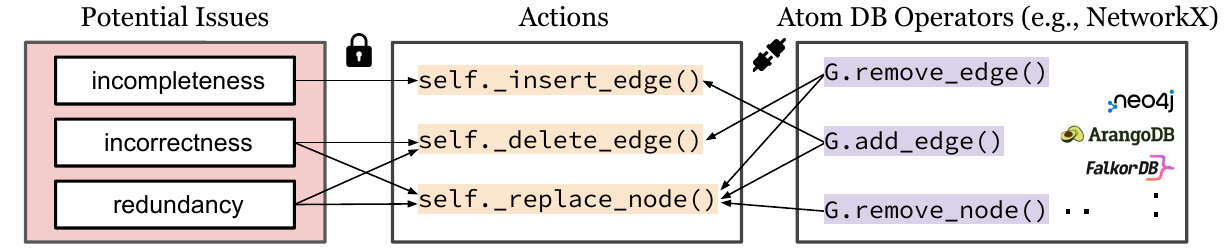}
   \caption{An overview about how the actions are corresponding to the potential issues in the KB and how the actions are executed with atom graph database operators in flexible database backends.}
   \label{fig:action}
 \end{figure}
 \vspace{-0.25cm}

We demonstrate how the three predefined types of actions are corresponding to the three aspects of potential issues in the KB and how the actions can be executed with atom graph database operators in flexible database backends in Figure~\ref{fig:action}. To be specific, from the functionality perspective, the edge insertion action complements missing relations or knowledge items to address incompleteness. Edge deletion action removes incorrect or redundant relations to mitigate errors and redundancy. And node replacement action resolves ambiguity to further reduce redundancy. Other actions are not included as they can either be implemented through the three pre-defined operators or are unnecessary (e.g., inserting an isolated node) for the knowledge refinement task. 

From the flexibility perspective, although the action space is fixed with the schema, they can be executed by different graph database backends via flexible mappings. The predefined actions are just interfaces between natural language sentences generated by the LLM and the common atom operators defined in various graph databases like NetworkX~\citep{hagberg2020networkx}, Neo4j, ArangoDB, and so on. For example, the node replacement action can be implemented through an integration of the original node deletion, original node related edge deletion, new node insertion and the edge insertion of the new node. Consequently, the fixed action space would not affect the generalizability across different backend graph database systems.

\section{Training System Overview of DeepRefine}
\begin{figure}[H]
  \centering
  \includegraphics[width=0.9\linewidth]{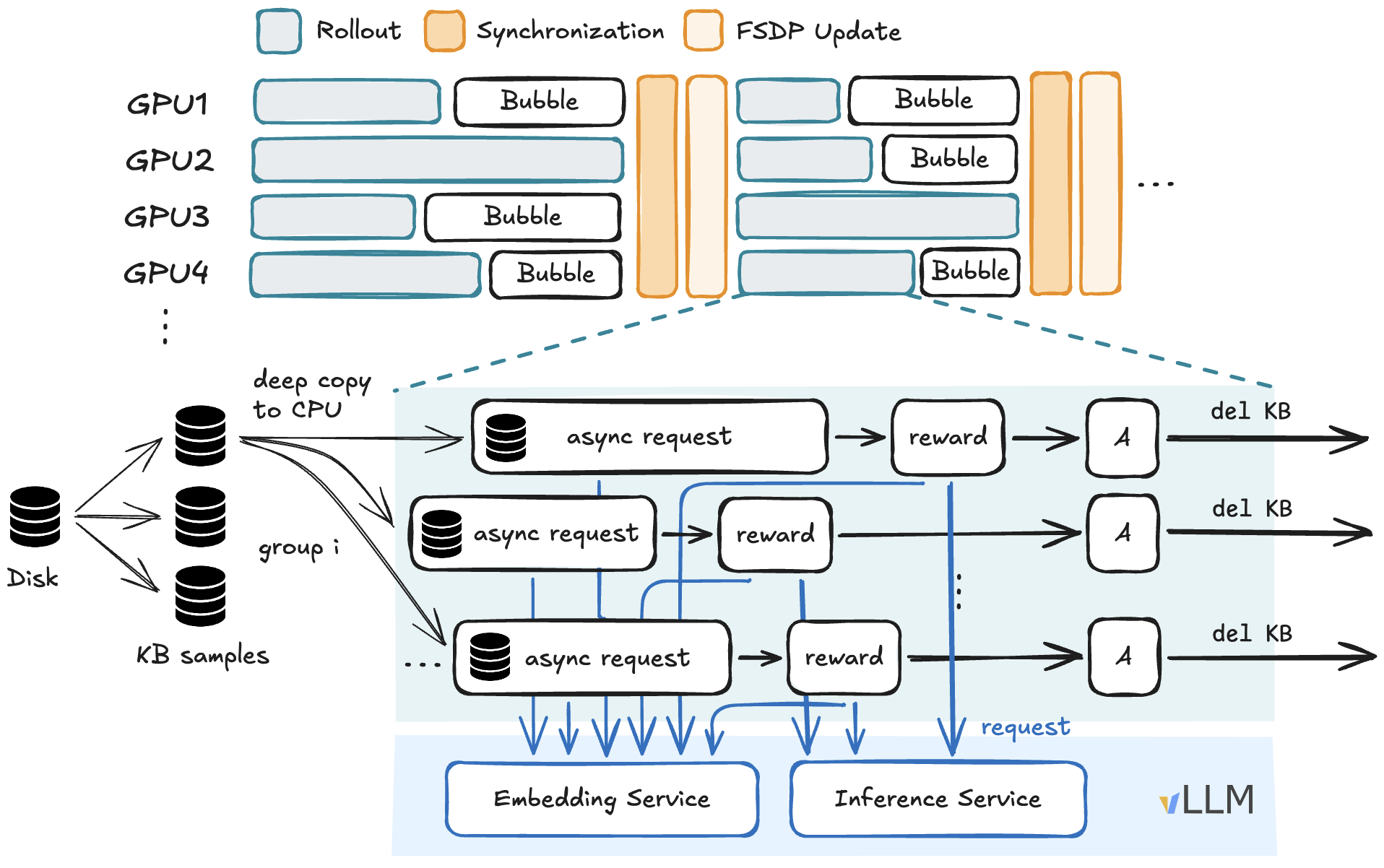}
  \caption{An overview about the training system of DeepRefine.}
  \label{fig:system}
\end{figure}
\vspace{-0.25cm}
To further understand the training workflow of DeepRefine and improve the reproductivity, we provide a detailed overview of the training system of DeepRefine in this section.

To be specific, as shown in the upper part of Figure~\ref{fig:system}, we adopt the verl~\citep{sheng2024hybridflow} framework to conduct the multi-turn RL training for its better systematic performance. In this figure, we take 4 GPUs and a group of samples as an example. Generally, the actor will update its policy when the slowest GPU is finished with the current batch of samples, the status of the other GPUs are all idle, and the synchronization across GPUs has also been finished.

Within each rollout in each GPU device, directly uploading a large knowledge base from the disk to the CPU memory and then make copies would lead to system crash. Because different rollouts should be conducted in isolated knowledge bases. So here we choose to use the small knowledge bases constructed from the local context of each query, which not only contains the query-related documents, but also the irrelevant documents. These small KB samples will first be sampled from disk and copy $G$ times to the CPU memory, which introduces much lower memory overhead. After each coroutine of the rollout and the reward computations are finished, the KBs will be remved from the CPU memory.

\section{Prompt Templates}
\label{sec:prompt_templates}
In this section, we will provide the prompt templates for the DeepRefine model. We demonstrate the prompt templates for each reasoning step in DeepRefine in Figure~\ref{answerable_judgement_prompt}, Figure~\ref{error_abduction_prompt}, and Figure~\ref{action_generation_prompt}.

\begin{figure}[H] 
  \begin{AIbox}{Prompt Template for Answerable Judgement.}
  {\bf System:}\\
  As an advanced judgement assistant, your task is to judge whether the given question is answerable based on the provided KG context.\\
  Evaluate whether the given question is answerable based on the provided KG context. Output your judgement in the following format:\\
{\color{opcolor}\texttt{<judge>}}Yes{\color{opcolor}\texttt{</judge>}} or {\color{opcolor}\texttt{<judge>}}No{\color{opcolor}\texttt{</judge>}}\\
**Important:** You must think carefully about the question and the KG context before making your judgement. And output your judgement result directly in the specified format.\\
\\
  {\bf User (first-turn):}\\
  Question: {\color{deepblue}\bf \{question\}} \\
  Knowledge Graph (KG) context: {\color{deepblue}\bf \{triples\_string\}} \\

  {\bf User (multi-turn)}:\\
  Additional KG context: {\color{deepblue}\bf \{triples\_string\}} \\
  Using the original question and all KG context in this conversation, judge answerability again.
  
  \tcblower
  {\bf \large Output:}
  \begin{lstlisting}[style=prompt]
  <judge>Yes</judge> or <judge>No</judge>
  \end{lstlisting}
  \end{AIbox}
  \caption{The prompt template for answerable judgement in DeepRefine.}
  \label{answerable_judgement_prompt}
\end{figure}

\begin{figure}[H] 
  \begin{AIbox}{Prompt Template for Error Abduction.}
  {\bf System:}\\
  As an advanced error abduction assistant, your task is to analyze the error reasons based on the given interaction history.\\
Analyze the reasons of the unanswerable questions based on the given interaction history. Output your analysis in the following format:\\
{\color{opcolor}\texttt{<abduction>}}...{\color{opcolor}\texttt{</abduction>}}\\
**Important:** You must think carefully about the interaction history before making your analysis. And output your analysis result directly in the specified format.\\
\\
  {\bf User:}\\
  The KG expansion stage is complete. Using the original question, all KG context, and judgement results already present in this conversation, analyze why the question was initially unanswerable.\\

  \tcblower
  {\bf \large Output:}
  \begin{lstlisting}[style=prompt]
  <abduction>...</abduction>
  \end{lstlisting}
  \end{AIbox}
  \caption{The prompt template for error abduction in DeepRefine.}
  \label{error_abduction_prompt}
\end{figure}

\begin{figure}[H] 
  \begin{AIbox}{Prompt Template for Refinement Action Generation.}
  {\bf System:}\\
  As an advanced knowledge graph refinement assistant, your task is to generate a series of actions to refine the given KG to make it more suitable for answering the given question.\\
Based on the given KG and the analysed error reasons, refine the given KG to make it more easily for retrieval and answering the given question. You have the following three types of actions to conduct:\\
- insert\_edge(subject, relation, object): Insert a new edge into the KG to complete the missing information.\\
- delete\_edge(subject, relation, object): Delete an edge from the KG to remove the redundant information or conflicting information.\\
- replace\_node(old\_entity, new\_entity): Replace an entity in the KG to correct the errors or deal with disambiguation.\\
Output a series of actions in the following format:\\
{\color{opcolor}\texttt{<refinement>}}insert\_edge("...", "...", "...")$|$delete\_edge("...", "...", "...")$|$replace\_node("...", "...")$|$...{\color{opcolor}\texttt{</refinement>}}\\
**Important:** You must think carefully about the given KG and the analysed error reasons before making your refinement. DO NOT DELETE ANY IRRELEVANT TRIPLES FROM THE ORIGINAL KG. TRY TO KEEP THE ORIGINAL KG AS MUCH AS POSSIBLE. And output your refinement result directly in the specified format.\\
\\
  {\bf User:}\\
Given the Original Text: {\color{deepblue}\bf \{original\_text\}}, \\
Using the original question, accumulated KG context, and the abduction immediately above, generate the graph refinement actions.
  \tcblower
  {\bf \large Output:}
  \begin{lstlisting}[style=prompt]
  <refinement>insert\_edge("...", "...", "...")|delete\_edge("...", "...", "...")|replace\_node("...", "...")|...</refinement>
  \end{lstlisting}
  \end{AIbox}
  \caption{The prompt template for refinement action generation in DeepRefine.}
  \label{action_generation_prompt}
\end{figure}

\section{Limitations}
\label{sec:limitations}
In this work, we propose an RL framework for agentic knowledge base refinement. Although DeepRefine has shown promising performance in the experiments, there are still some potential improvements that can be made to the model. For example, we only construct the training data from HotpotQA with easy filtering criteria, which makes the quality of the training data not high enough. Finding out higher quality and more diverse training data is a promising direction for future work. And how to combine DeepRefine with a robust systematic refinement framework to make the performance better and more stable is also underexplored.

\end{document}